%% file: acl_latex.tex
\pdfoutput=1

\documentclass[11pt]{article}

\usepackage[]{acl}

\usepackage{times}
\usepackage{latexsym}

\usepackage[T1]{fontenc}

\usepackage[utf8]{inputenc}

\usepackage{microtype}

\usepackage{enumitem}
\usepackage{graphicx}
\usepackage{booktabs}
\usepackage{multirow}
\usepackage{longtable}
\usepackage{amsmath}
\usepackage{cleveref}
\usepackage{url}
\usepackage{fancyvrb}
\usepackage[]{todonotes}
\setuptodonotes{inline}
\usepackage{tabularx} 
\usepackage{array}    
\usepackage{ragged2e} 
\usepackage{float}
\usepackage{subcaption}

\usepackage{xcolor}             
\usepackage[most]{tcolorbox}    

\newcommand{\fitcol}[2]{\begin{minipage}{#1\textwidth}#2\vspace{0.5em}\end{minipage}}

\usepackage{listings}
\usepackage{inconsolata} 
\newcommand{\plexmono}{\ttfamily}
\definecolor{boxframecolor}{gray}{0.75}
\definecolor{boxtitlecolor}{gray}{0.85}
\newtcolorbox{promptbox}[1]{
    enhanced,
    width=\textwidth,               
    arc=1mm,                        
    boxrule=0.8pt,                  
    colback=white,                  
    colframe=boxframecolor,         
    colbacktitle=boxtitlecolor,     
    coltitle=black,                 
    fonttitle=\bfseries\sffamily,   
    title=#1,                       
    before upper={%
        \plexmono 
    },
}

\definecolor{boxbackcolor}{gray}{0.95} 

\title{Can one size fit all?: Measuring Failure in Multi-Document Summarization Domain Transfer}

\author{Alexandra DeLucia
    \and
    Mark Dredze \\
    Center for Language and Speech Processing\\
    Johns Hopkins University\\
    \texttt{\{aadelucia,mdredze\}@jhu.edu}
}

\begin{document}
\maketitle

\begin{abstract}
Abstractive multi-document summarization (MDS) is the task of automatically summarizing information in multiple documents, from news articles to conversations with multiple speakers. 
The training approaches for current MDS models can be grouped into four approaches: end-to-end with special pre-training (``direct''), chunk-then-summarize, extract-then-summarize, and inference with GPT-style models.
In this work, we evaluate MDS models across training approaches, domains, and dimensions (reference similarity, quality, and factuality), to analyze \textit{how} and \textit{why} models trained on one domain can fail to summarize documents from another (News, Science, and Conversation) in the zero-shot domain transfer setting.
We define domain-transfer ``failure'' as a decrease in factuality, higher deviation from the target, and a general decrease in summary quality.
In addition to exploring domain transfer for MDS models, we critically examine the behavior of various automatic evaluation metrics, including their inter-correlations and the performance of LLM-as-a-judge approaches, in these transfer settings.
\end{abstract}

\section{Introduction}
\label{sec:intro}

Multi-document summarization (MDS) is an extended task of summarization, where the goal is to represent the key ideas from multiple documents in a single new document or paragraph.
Current MDS systems (models) are trained and evaluated on data from a single domain, such as news, but there have been no robust studies on how models trained on these domains transfer, or perform, on different domains.
While the overarching goal of MDS models is to condense information, models trained on one type of summary (e.g., news) often fail to transfer successfully to other domains (e.g., science) due to the unique nature of how summaries across different domains are written.

\input{fig1}

Recent MDS system surveys compare different models either on a single domain (e.g., conversation/dialogue \citep{zhang-etal-2021-exploratory-study,mullick2024long} or focus on MDS corpora biases \citep{wolhandler-etal-2022-multi,dey-etal-2020-corpora}.
While \citet{zhang-etal-2021-exploratory-study} found that retrieve-then-summarize models are best according to ROUGE, \citet{mullick2024long} did not find a clear pattern in model performance across a variety of metrics.
Few works explore the zero-shot domain transfer ability of MDS systems, but \citet{zhang-etal-2021-exploratory-study} did find that pre-training on news datasets can transfer well to conversation. However, \citet{fabbri-etal-2021-convosumm} a zero-shot model from news to conversation had lower performance than on conversation alone.

An example is in \Cref{fig:transfer_ex}, where two models, one trained on Multi-News+ \citep{choi-etal-2024-multi-news,fabbri-etal-2019-multi} and the other on Multi-XScience \citep{lu-etal-2020-multi-xscience} are evaluated on a sample from ConvoSumm-Reddit \citep{fabbri-etal-2021-convosumm}.
The difference in writing style is immediately apparent: the news-trained model has a blogger-like style, the science-trained model inserts citations, and both are completely different from the expected style of the Reddit conversation summary.
The difference in style is also captured by similarity metrics ROUGE \citep{lin-2004-rouge} and UniEval-Relevance \citep{zhong-etal-2022-towards}--while the ROUGE scores, which capture word overlap, only differ by a few points, the semantic-based, trained Relevance scores vary widely. 

While ConvoSumm-Reddit summaries are written to represent multiple points of view, other ways to represent online discourse is through an overall topic-based summary or detailed utterance-by-utterance summary.
All types would ``condense'' the same conversation but at varying levels of granularity, style, and word choice.
Similarly, summaries for other domains also vary in incorporated information and style.

\Cref{fig:transfer_ex} highlights the three axes of domain transfer for MDS systems: 1) domain of the training data, 2) type of model, and 3) evaluation metrics.
While there are popular benchmarks for summarization, including the notable SummEval \citep{fabbri-etal-2021-summeval}, few are for multi-document summarization, and most datasets focus on the news domain.
We choose three high-quality human-written datasets to represent News (Multi-News \citep{fabbri-etal-2019-multi}), Science (Multi-XScience \citep{lu-etal-2020-multi-xscience}), and Conversation (ConvoSumm-Reddit \citep{fabbri-etal-2021-convosumm}) domains. 

Regarding the type of model, while there are many models for summarization, the types of models for \textit{abstract} multi-document summarization are end-to-end with special pre-training (``direct''), chunk-then-summarize, extract-then-summarize, and direct with large language model (LLM) decoders. 
There are many models within each type \citep{mullick2024long} and for greater in-depth analyses, we choose one, strong representative model of each kind.
We focus on neural, transformer models and omit extractive comparisons due to previous work demonstrating the stronger performance of neural models \citep{fabbri-etal-2021-summeval}.

Similarly, while there are many evaluation metrics for generated text, which each have their strengths and weaknesses \citep{he-etal-2023-blind, von_daniken_effectiveness_2022,fabbri-etal-2021-summeval}, we group commonly used metrics into three categories: 1) similarity to summary reference (ground-truth), 2) overall quality (e.g., fluency), and 3) groundedness in the source documents. We find that:\footnote{All code and models will be released upon publication.}

\begin{itemize}[noitemsep]
    \item LLMs (e.g., Llama 3.1) demonstrate robust quality and factuality in domain transfer with minimal performance change, and their outputs are less influenced by few-shot examples compared to the stronger stylistic bias observed in smaller, fine-tuned models;
    \item While no single source domain (News, Science, or Conversation) proves universally superior for general transfer, the stylistic nature of summaries learned during training significantly impacts cross-domain performance;
    \item Lexical similarity metrics (e.g., ROUGE, MoverScore) show weak correlation with semantic relevance (e.g., UniEval-Relevance) and other quality dimensions; 
    \item Source-dependent non-reference metrics face scaling challenges moving from single-document summarization to multi-;
    \item Trained multi-dimensional evaluators exhibit intra- and inter-correlations with each other and within dimensions;
    \item The number of source documents surprisingly shows weak and inconsistent correlations with factuality and coherence metrics in domain-transfer settings across different model architectures;
\end{itemize}

\begin{figure*}[t]
    \centering
    \includegraphics[width=\textwidth]{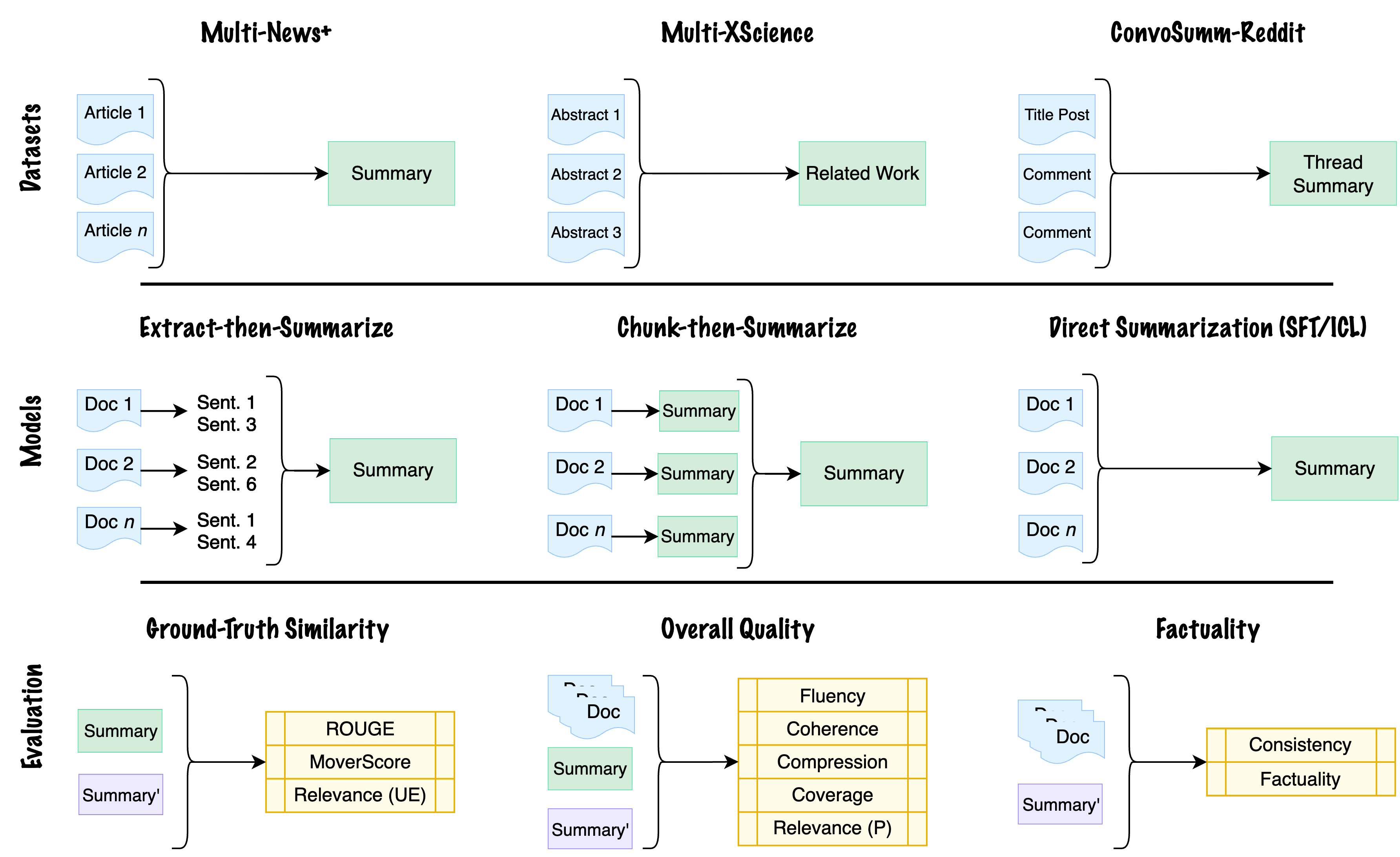}
    \caption{Summary of all the domains (datasets), multi-document summarization (MDS) model types, and evaluation metrics analyzed in this work. The ``Relevance'' metric is in both Ground-Truth Similarity and Overall Quality groups because the UniEval (UE) and Prometheus (P) definitions are different (see \Cref{fig:prometheus_criteria,fig:unieval_criteria} for details).}
    \label{fig:experimental-summary}
\end{figure*}

\section{Domain Transfer of Multi-Document Summarization}
\label{sec:related_work}

While there is in-depth work on analyzing summarization and MDS systems and corpora \citep{zhang-etal-2021-exploratory-study,mullick2024long,fabbri-etal-2021-summeval,dey-etal-2020-corpora}, there is no systematic evaluation of how each model type performs when applied out-of-domain and how the automatic evaluation metrics can over- or under-represent model performance. 
We draw from existing models and tools to explore the intricacies of domain transfer with MDS models and their evaluation methods.
The main dimensions of our analyses are domain (i.e., dataset), model type, and evaluation metrics. See \Cref{fig:experimental-summary} for summary.

\subsection{Summarization Dataset Domains}
News is a common domain for summarization datasets \citep{hermann_teaching_2015,fabbri-etal-2019-multi,jiang_ccsum_2024,choi-etal-2024-multi-news,gholipour-ghalandari-etal-2020-large,litkowski_summarization_2004}
There are datasets for other domains like scientific articles \citep{lu-etal-2020-multi-xscience} and conversation \citep{fabbri-etal-2021-convosumm,gliwa-etal-2019-samsum,chen-etal-2021-dialogsum-challenge}. 
We select a single, high-quality multi-document dataset with human-written summaries to represent each domain. All documents and summaries are in English.

\paragraph{Conversation.}
We use the Reddit subset of the released data from ConvoSumm \citep{fabbri-etal-2021-convosumm} (\textbf{ConvoSumm-Reddit}). Since the summaries were written by paid crowdsource workers, this dataset is significantly smaller than the news and science datasets, with only 500 entries.

\paragraph{News.} \textbf{Multi-News+} \citep{choi-etal-2024-multi-news}, an automatically cleaned version of Multi-News \citep{fabbri-etal-2019-multi}, is a news-based MDS dataset featuring human-written summaries from Newser.\footnote{\url{newser.com}}. Multi-News+ contains 44,668/5,585/5,584 entries in the train/validation/test sets.

\paragraph{Science.}
Representing the Science domain, the entries in \textbf{Multi-XScience} pair a related works section of an academic paper with 1) its abstract and 2) the abstracts from cited work. The 40,528 papers were collected from arXiv and the Microsoft Academic Graph (MAG) \citep{lu-etal-2020-multi-xscience}. Since the paper dump was from pre-2020, we assume all writing on arXiv is human-generated.

\subsection{Multi-Document Summarization Models}
Current high-performing fine-tuned abstractive summarization methods can be grouped into four approaches: summary-specific pre-training \citep{zhang_pegasus_2020,xiao-etal-2022-primera,peper_pelms_2024}, incremental summarization (``chunk-then-summarize'') \citep{zhang-etal-2022-summn}, extracting claims or viewpoints (``extract-then-summarize'') \citep{fabbri-etal-2021-convosumm, ernst-etal-2022-proposition, ouyang-etal-2023-compositional}, and fine-tuning or few-shot summarization with large pre-trained models \citep{mullick2024long}. 
These groupings have been used in prior work \citep{mullick2024long}.

\paragraph{Modified Pretraining.} 
Prior work has shown that fine-tuning a language model specifically pre-trained with summary-focused objectives improves downstream performance over non-specialized models \citep{xiao-etal-2022-primera,zhang_pegasus_2020,peper_pelms_2024}.
\citet{zhang_pegasus_2020} first introduced this concept of summary-specific pre-training, and it was later augmented to the multi-document case \citep{xiao-etal-2022-primera,peper_pelms_2024}.
We select \textbf{PRIMERA} to represent this group. PRIMERA is Longformer-Encoder-Decoder model initialized from BART-Large \citep{beltagy_longformer_2020} and pre-trained with an Entity Pyramid objective on a large news corpora \citep{xiao-etal-2022-primera}. 
While PELMS \citep{peper_pelms_2024} adjusts the pre-training strategy to improve upon PRIMERA, we omit comparison to this model because it also uses a data filtering step, which overlaps with the extract-then-summarize approach.

\paragraph{Chunk-then-summarize.} Another approach to MDS is incremental summarization of documents, as done with \textbf{Summ$^N$} \citep{zhang-etal-2022-summn}. 
Summ$^N$ first segments the source documents and target summary and creates mappings between the source and target segments (via ROUGE max similarity), which and trains a separate BART-Large-CNN model \citep{zhang-etal-2021-exploratory-study} each ``coarse'' stages before the final ``fine-grained'' stage which produces the end summary.

\paragraph{Extract-then-summarize.} 
A different approach to the long-context problem of MDS systems is to first \textit{extract} the salient portions of the documents before summarizing them. 
The main models in this category are COMPO \citep{ouyang-etal-2023-compositional}, ProCluster \citep{ernst-etal-2022-proposition}, and ConvoSumm \citep{fabbri-etal-2021-convosumm},. 
COMPO is specifically tailored toward conversational summarization and provides summaries based on extracted sub-components (topics and actions) within the dialogue.
ProCluster extracts ``propositions'' in documents, clusters them, and then summarizes the clusters.
ConvoSumm first filters sentences from documents to only the ``claims'' and then summarizes the claims.
A variation of the extract-then-summarize pipeline is attribute-then-generate \citep{slobodkin_attribute_2024} which requires multiple steps of content selection, sentence planning, and sentence-by-sentence generation.
We chose \textbf{ConvoSumm} over other extract-then-summarize models due to its applicability to many domains and non-overlap with different approaches with clustering or sentence-planning steps.

\paragraph{Pre-trained LLM.} 
While the above models are strong MDS systems, they are all the size of BART-Large (336M params.).
A growing trend across natural language processing research has been to fine-tune or adapt large, pre-trained decoder language models (LLMs) through in-context learning.
We select \textbf{Llama-3.1-8B-Instruct} to represent this approach due to its strong performance on a variety of tasks \citep{dubey_llama_2024}.

\subsection{Summarization Evaluation}
Most MDS papers and their surveys use off-the-shelf automatic evaluation metrics without discussing drawbacks. 
Summarization metrics usually measure one or more of the following attributes: similarity to a ground-truth summary, overall quality, and factuality.
While many MDS systems are evaluated with manual annotations, we focus on automatic evaluation in this work.

\paragraph{Similarity to Ground Truth.}
Most text evaluation metrics compare the model output to a known, quality summary or \textit{ground truth} (i.e., reference).
The most common metric for evaluating summary quality is \textbf{ROUGE} \citep{lin-2004-rouge} but is better viewed as a \textit{similarity} metric since it is the token recall of the generated summary based on the reference (i.e., ground-truth). 
The complement of ROUGE is BLEU \citep{papineni-etal-2002-bleu}, which measures precision instead of recall of generated summaries.

Other metrics in the spirit of ROUGE/BLEU are semantic-based instead of lexical, which are more robust to passages with similar meanings but different word choices. 
The most popular version is BERTScore \citep{zhang_bertscore_2020}, which maps tokens from the generation to the reference before calculating overlap.
An alternative is \textbf{MoverScore}, an ``optimized'' version of BERTScore which uses soft token alignments from the hypothesis to the reference instead of a hard 1:1 mapping, allowing for a greater reward for semantically similar generated text.

Other similarity metrics are sub-``dimensions'' of a trained multi-dimensional evaluator, like UniEval \citep{zhong-etal-2022-towards}. 
UniEval is a T5-Large model fine-tuned on synthetically augmented datasets for positive and negative examples of different ``dimensions'' of summary attributes (fluency, relevance, coherence, consistency) \citep{zhong-etal-2022-towards}. 
The \textbf{UniEval-Relevance} dimension is a similarity metric that compares the reference and generation as a whole instead of per token.

\paragraph{Overall Quality.}
Not all desired summary qualities depend on a reference; some are intrinsic or rely on the source documents. Intrinsic metrics include \textit{fluency} (i.e., well-written and grammatical), \textit{coherence} (i.e., overall flow), and the summarization-specific metrics of  \textit{compression ratio} (i.e., length of source documents compared to length of the summary) and \textit{source coverage} (i.e., how well the summary represents the source documents).
UniEval measures both fluency and coherence.
\textbf{UniEval-Fluency} is based solely on the generation, and \textbf{UniEval-Coherence} measures the coherence of the generation with respect to the source.

Similar to UniEval, Prometheus \citep{kim-etal-2024-prometheus} is another multi-dimensional evaluator, but is a general-purpose feedback model that can judge any model output according to a provided rubric. In this respect, Prometheus is essentially a distilled ``LLM-as-a-Judge'' model \citep{zheng_judging_2023}. For a more up-to-date automatic evaluation method, we use Prometheus to grade the summaries on the following criteria, drawn from SummEval \citep{fabbri-etal-2021-summeval}: Relevance, Coherence, and Fluency. An important note is that despite the same metric names, the definitions from SummEval differ slightly from those in UniEval.

To measure compression ratio, we include both \textbf{Compression-Sentence}, which compares the ratio of the number of sentences in the summary to those in the source documents, and \textbf{Compression-Word}, which compares the ratio of the number of words in the summary to those in the source documents \citep{koh_empirical_2022}. For source coverage \citep{wolhandler-etal-2022-multi}, we developed a prompt for Prometheus, \textbf{Prometheus-Coverage}.

\paragraph{Factuality.}
A growing concern with generated text is the possibility of hallucinations and non-faithfulness to the source material \citep{min_factscore_2023}. 
While objective factuality is important for the safety of LLMs (e.g., knowing that boiling water is hot), for MDS we typically assume that the source documents are factually correct and that a summary is only factual if its claims match the claims from the source. This definition of ``factually'' is better described as \textit{grounding in the source text}.
We measure summary factuality with \textbf{UniEval-Consistency}, a single-dimensional evaluator trained to measure the factual consistency of generated text to the source (as opposed to continuously trained as with UniEval-Fluency and UniEval-Coherence), \textbf{Prometheus-Consistency}, and \textbf{Prometheus-Factuality}. The Prometheus-Consistency prompt is drawn from SummEval, and the Prometheus-Factuality prompt is from the Prometheus repository.

We include \textbf{ROUGE} ($[0,1]$), \textbf{MoverScore} ($[-1,1]$), \textbf{Compression Ratio}, and multiple dimensions of \textbf{UniEval} ($[0,1]$) and \textbf{Prometheus} ($\{1, 2, 3, 4, 5\}$) in our analysis as representations for similarity to ground truth, factuality, and overall quality metrics. 
While BARTScore \citep{yuan2021bartscore} measures many of the same dimensions as UniEval, we chose UniEval because it is better correlated with human evaluation \citep{zhong-etal-2022-towards} and the potential bias of BARTScore to model generations, since most of the models are BART-based (``self bias'') \citep{he-etal-2023-blind}.

\section{Experimental Setup}
\label{sec:setup}
Our goal is to explore the intricacies of domain transfer with multi-document summarization models and their evaluation methods.
The main dimensions of our analyses are different datasets (i.e., domains), models, and evaluation metrics (\Cref{fig:experimental-summary}). 
We chose 3 dataset domains (News, Science, Conversation), 4 model types (end-to-end/direct), chunk-then-summarize, extract-then-summarize, and pre-trained LLMs), and 15 metrics for a total of 180 points of analysis (\Cref{sec:related_work}).

\subsection{Model Training}
The hyperparameters for each model and dataset combination are in  \Cref{app:train_details}.
Each model was trained on each dataset, and we refer to them as \texttt{ModelName}-\texttt{Domain}, e.g., ConvoSumm fine-tuned on Multi-News+ is ConvoSumm-News.

\paragraph{PRIMERA.} As input, the source documents are delineated with a special \texttt{<doc-sep>} token, and the model's global attention is set on those tokens.
To handle examples with source documents that are too long for the context window, we follow the author's code and evenly truncate the documents to fit.
We used the authors' released PRIMERA-MultiXScience model and trained PRIMERA-Convo and PRIMERA-News with the released code.\footnote{While there is a PRIMERA model trained on Multi-News, we trained a new one on Multi-News+.}

\paragraph{Summ$^N$.} We use a 3 course-stage process for all datasets.
A separate model is trained for each coarse stage, leading to $N + 1$, or $4$ fine-tuned BART models for each training dataset.
An important note is due to the setup of summarizing chunks, there are no too-long context issues as with direct models like PRIMERA and all source documents are considered by the model. 

\paragraph{ConvoSumm.}
There are multiple versions of ConvoSumm but we select ConvoSumm-argument-filtered (``arg-filtered'') due to its comparable performance to the more complex argument-graph version.
The argument filtration model is a BERT model fine-tuned on annotated data from the Change My View community on Reddit (i.e., conversation, AMPERSAND \citep{chakrabarty-etal-2019-ampersand}) and annotated persuasive essays on a variety of topics \citep{stab-gurevych-2017-parsing}). A BART-Large model with an expanded context window (2048) is fine-tuned on the extracted claims.

\paragraph{Llama 3.1.}
Unlike the other models, we do not fine-tune Llama 3.1 8B-Instruct.\footnote{\url{https://huggingface.co/meta-llama/Llama-3.1-8B-Instruct}}
To compare to the domain-transfer experiments, we provide an example from the training set to the model, i.e., for News-Convo, two randomly selected examples from Multi-News+ is provided to Llama and then the model is prompted to summarize an example from ConvoSumm. 
We also include a zero-shot version which does not provide a complete example to Llama, only the instruction.

\subsection{Evaluation Details}
ROUGE scores have been known to vary depending on the implementation used. We use the HuggingFace wrapper for the Google Research implementation.\footnote{\url{https://huggingface.co/spaces/evaluate-metric/rouge}}
We report \textbf{ROUGE-1} (R1) and \textbf{ROUGE-L-Sum} (RL-Sum) to follow prior work.
For the other evaluation metrics, \textbf{MoverScore} and multiple dimensions from \textbf{UniEval}, we use the trained models as they are without modification.
For evaluation with \textbf{Prometheus}, we use Prometheus2-BGB-8x7B \footnote{\url{https://huggingface.co/prometheus-eval/prometheus-bgb-8x7b-v2.0}} as an absolute judge with dimension-specific rubrics we based off the definitions from SummEval \citep{fabbri-etal-2021-summeval}. See \Cref{app:evaluation} for details. Due to the intense compute required for Prometheus, we sampled 100 items from each dataset for all Prometheus-based metrics.

\begin{table*}
\resizebox{\textwidth}{!}{%
\begin{tabular}{@{}p{0.15\textwidth} l r r | r r | r r | r r | r@{}} 
\toprule
& & \multicolumn{2}{c}{ConvoSumm} & \multicolumn{2}{c}{PRIMERA} & \multicolumn{2}{c}{Summ$^N$} & \multicolumn{2}{c}{Llama 3.1} & \multicolumn{1}{c}{\begin{tabular}{@{}c@{}}Llama 3.1 \\ (0-shot)\end{tabular}} \\
&  & \multicolumn{1}{c}{} & \multicolumn{1}{c}{Avg. $\Delta$} & \multicolumn{1}{c}{} & \multicolumn{1}{c}{Avg. $\Delta$} & \multicolumn{1}{c}{} & \multicolumn{1}{c}{Avg. $\Delta$} & \multicolumn{1}{c}{} & \multicolumn{1}{c}{Avg. $\Delta$} & \multicolumn{1}{c}{Avg. $\Delta$} \\
\midrule
\multirow{3}{=}{Factuality} & Consistency (UE) & 0.67 & -2\% & 0.82 & 0\% & 0.65 & \textbf{+3\%} & 0.86 & 0\% & 0\% \\ 
& Factuality (P)    & 2.04 & -16\% & 1.79 & \textbf{+3\%} & 2.02 & -6\% & 3.23 & 0\% & 0\% \\ 
& Consistency (P)   & 2.14 & -15\% & 1.89 & \textbf{+1\%} & 2.08 & -7\% & 3.40 & +1\% & 0\% \\ 
\midrule
\multirow{4}{=}{Ground-Truth Similarity} & Relevancy (UE) & 0.66 & -3\% & 0.48 & \textbf{+6\%} & 0.66 & -19\% & 0.92 & 0\% & 0\% \\ 
& ROUGE-1         & 0.35 & -27\% & 0.27 & \textbf{+12\%} & 0.35 & -30\% & 0.29 & +10\% & 0\% \\ 
& ROUGE-LSum      & 0.20 & -28\% & 0.14 & \textbf{+6\%} & 0.20 & -28\% & 0.17 & +4\% & +1\% \\ 
& MoverScore          & 0.10 & -86\% & 0.06 & -332\% & 0.10 & -88\% & 0.10 & \textbf{+35\%} & -2\% \\ 
\midrule
\multirow{8}{=}{Quality} & Relevancy (P)   & 2.36 & -15\% & 2.18 & \textbf{+3\%} & 2.33 & -6\% & 3.53 & +1\% & +1\% \\ 
& Fluency (UE) & 0.79 & -5\% & 0.69 & \textbf{+2\%} & 0.76 & 0\% & 0.94 & 0\% & 0\% \\ 
& Fluency (P)   & 2.30 & -21\% & 1.67 & \textbf{+9\%} & 2.26 & -5\% & 3.97 & 0\% & 0\% \\ 
& Coherence (UE) & 0.60 & -14\% & 0.70 & \textbf{+2\%} & 0.63 & -2\% & 0.95 & 0\% & 0\% \\ 
& Coherence (P)   & 2.31 & -20\% & 1.63 & \textbf{+8\%} & 2.31 & -11\% & 3.95 & 0\% & 0\% \\ 
& Comp.-Sent    & 0.71 & +7\% & 0.66 & +1\% & 0.69 & \textbf{+9\%} & 0.64 & +4\% & 0\% \\ 
& Comp.-Word    & 0.74 & +1\% & 0.55 & \textbf{+2\%} & 0.70 & -1\% & 0.55 & +2\% & 0\% \\ 
& Coverage (P)  & 1.93 & -12\% & 1.83 & \textbf{+3\%} & 1.95 & -6\% & 3.03 & +1\% & 0\% \\ 
\bottomrule
\end{tabular}
}
\caption{Multi-document summarization (MDS) model performance and relative performance \textbf{aggregated by model} when trained and tested on the same domain (left) and different domains (right, $\Delta$), respectively. A relative change of over $1$ is possible with MoverScore because of its $[-1,1]$ range. Llama (0-shot) is compared to Llama with in-domain examples.}
\label{tab:results-model}
\end{table*}

\section{Results}
\label{sec:results}
We analyze the results in the framework of our three axes: dataset domain, model type, and evaluation metric.\footnote{The domain-transfer analysis is in \Cref{app:results-domain}.}
See \Cref{app:results-raw} for the detailed per-model and dataset results.
To study evaluation metrics, we analyze the performance of the ground-truth dataset.
Complete summary examples from each model are in Appendix \Cref{fig:convosumm_summaries,fig:icl_summaries,fig:sumn_summaries,fig:primera_summaries,fig:zeroshot_summary}.

\subsection{MDS Model Generalizability}
\label{sec:results-transfer}
To study trends specific to each model type, we aggregated the metrics across all the datasets (\Cref{tab:results-model}). We also look at correlations between metrics in Appendix \Cref{fig:corr_model}.
Instead of judging performance by the raw metrics, we look at the \textit{relative change in performance} from the in-domain setting to the out-of-domain setting (e.g., training on ConvoSumm-Reddit and testing on Multi-XScience). 
The zero-shot baseline with Llama 3.1 is compared to the in-context Llama 3.1 performance for each dataset.
A positive $\Delta$ shows a performance gain and a negative $\Delta$ is a performance loss as compared to the model's scores on the in-domain data. 

\paragraph{Llama 3.1 outperforms all fine-tuned smaller models in quality and factuality metrics.}
Across all datasets in the in-domain setting, Llama generates the most factual and highest-quality summaries according to the automatic metrics.
This trend holds during domain transfer, with minimal change in performance ($+1$ to $+4$).
The exception to the higher performance trend is with the compression metrics and lexical-similarity metrics. Llama has the lowest compression, similar to PRIMERA, as summaries are only $55\%$ compressed. 

\paragraph{Zero-shot performance is the same as in-context learning with Llama 3.1.}
Surprisingly, Llama has a similar performance at generating summaries with (in-context learning, ICL) or without being shown an example (zero-shot).
The Llama-generated summaries change very little when shown examples versus when no examples are provided (Appendix \Cref{fig:zeroshot_summary,fig:icl_summaries}).
This indicates a strong bias to its pre-training versus ``learning'' from the example.
These results could differ with the addition of more examples (at the expense of a smaller context window available for the source documents) or a larger model. 

\paragraph{Not all domains benefit from Extract-then-Summarize models.}
The first step in the extract-then-summarize model ConvoSumm is to filter out sentences that are not identified as claims by a small, fine-tuned model.
This step removes $35\%$ of sentences on average from each ConvoSumm-Reddit example, $51\%$ from Multi-News, but less than $1\%$ from Multi-XScience.
This is intuitive since there would be more filler sentences in conversations and news articles, but less so in scientific publications. 

\paragraph{Chunk-then-summarize is the only approach guaranteed to fit all source documents.}
With all models, including Llama 3.1 with a 131K context window, there were issues fitting all source documents for every example comfortably in the context windows.
As mentioned in \Cref{sec:setup}, PRIMERA ``handles'' this issue by evenly dividing the context window among the source documents and truncating the ends. 
The assumption that the beginning of documents is more important than the end does not hold for all domains and has only been identified in News \citep{dey-etal-2020-corpora}.
While ConvoSumm was able to fit all source documents from ConvoSumm-Reddit, it did have to truncate documents from 25 samples in Multi-XScience and 699 samples from MultiNews.
PRIMERA, with a context window double the size of ConvoSumm (2K vs 4K), was able to fit all Conversation and Science samples, but truncated 835 Multi-News examples. 
Llama only had issues with 3 samples from Multi-News.

\paragraph{The number of source documents does not have strong correlations with other metrics.} Surprisingly, the number of source documents is weakly and inconsistently correlated with factuality and coherence metrics ($-0.1$ to $-0.5$). This is different than expected, as model performance is expected to decrease with longer context and more information to process.

\paragraph{All models are fluent.} Unsurprisingly, all models have high fluency scores ($0.70+$) since all models are of size BART-Large and above. While fluency is still an important attribute to check, it might not be as important as it once was.

\paragraph{Word-based similarity metrics are not correlated with Relevancy.}
While UniEval-Relevancy measures the similarity between a generated and reference summary (``Is this summary relevant to the reference?''), it does not correlate with the semantic and lexical metrics like MoverScore and ROUGE ($-0.1$ to $0.1$). Whereas MoverScore and ROUGE are highly correlated ($0.6$ to $0.8$).

\subsection{Evaluation of Evaluation}
\label{sec:analysis-metrics}
\begin{figure}
    \centering
    \includegraphics[width=\linewidth]{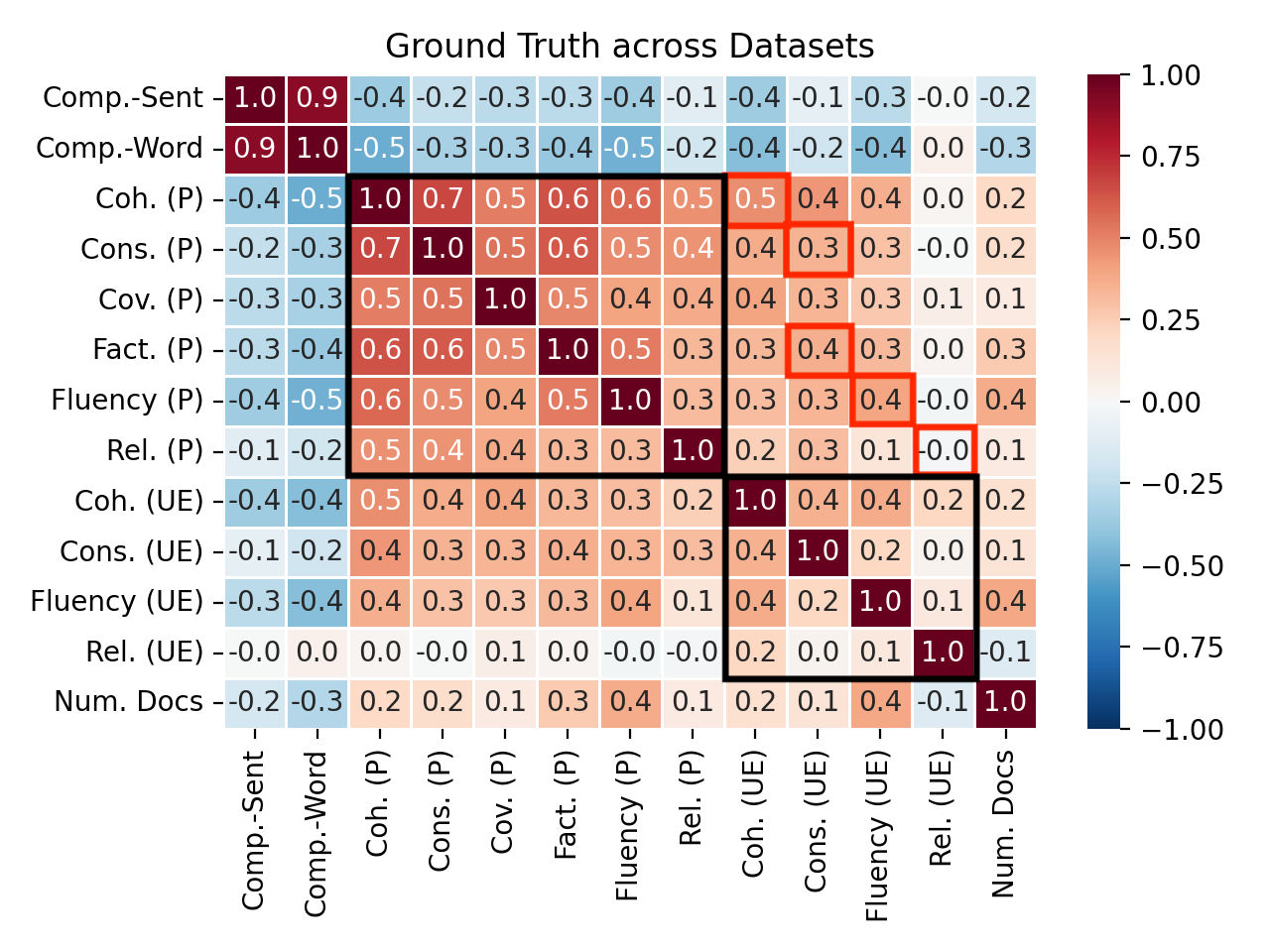}
    \caption{Pearson correlations across the ground-truth for all evaluated datasets. The black squares section off intra-correlations within Prometheus and UniEval, and the red square highlights the correlations between them.}
    \label{fig:corr-data-summ}
\end{figure}

To isolate patterns of the evaluation metrics, we look at the automated metric values on the ground-truth reference datasets (Appendix \Cref{tab:results-data}) and their correlations (\Cref{fig:corr-data-summ}). See \Cref{app:results-raw} for detailed correlation figures.


\paragraph{UniEval and Prometheus have a weak to moderate correlation.}
Over the three datasets, the UniEval and Prometheus metrics have weak to moderate correlations ($0.3$ to $0.5$). While Prometheus is $60x$ larger than UniEval (T5-Large), a stronger positive correlation is expected when they are evaluating the same quality (e.g., fluency). The extreme outlier is UniEval-Relevance, which has poor to no correlation with any other metric ($0.0$ to $0.2$).

\paragraph{Metric names are not consistent across evaluators and benchmarks.}
The definition of ``relevance'' differs between UniEval and SummEval, ``Is this summary relevant to the reference?'' versus ``How well does the summary capture the key points of the documents?''. This discrepancy explains the $0.0$ correlation between UniEval-Relevance and Prometheus-Relevance. The other metrics have similar definitions and are more correlated, as noted above. See Appendix \Cref{fig:prometheus_criteria,fig:unieval_criteria} for dimension definitions.

\paragraph{Multi-dimensional evaluators are slightly intra-correlated.}
Within Prometheus, the dimensions have slight ($0.4$) to moderate ($0.7$) correlations with each other, specifically coherence, consistency, and factuality ($0.6$ to $0.7$). The UniEval dimensions are less intra-correlated, ($0.0$ to $0.4$). This discrepancy in intra-correlation could indicate that Prometheus has difficulty evaluating along a single dimension and instead allows various quality aspects to impact the assigned score. This issue persisted despite adding instructions only to evaluate the desired dimension (see \Cref{app:evaluation}).

\paragraph{Non-reference metrics do not scale well to the multi-document setting.}
Evaluating for attributes that depend on the source documents (e.g., Consistency) is most affected by the long-document nature of multi-document summarization. 
The options are either to 1) truncate the documents, which can greatly alter the scores depending on which documents are included, or 2) calculate the metric for \textit{every} source document and aggregate the scores.
More work is needed to determine whether truncating or aggregating per-document scores over- or under-estimates model performance.
As for the now-popular ``LLM-as-a-Judge'' models, like Prometheus, while they have less of a context-window issue, they have a higher computational cost.
The reference-based metrics (primarily similarity-based) are not affected by scale because they only compare the ground-truth and generated summaries.

\section{Discussion and Conclusion}
\label{sec:conclusion}

In most work that surveys or proposes multi-document summarization (MDS) systems, there is little overlap between the analysis of model performance and the shortcomings of popularly used summarization evaluation metrics. We bridged this gap through experiments and analysis of not only the zero-shot domain transfer ability of MDS systems but also a critical examination of their evaluation methodologies. We analyzed three components of the MDS pipeline for domain transfer: 1) the domain of the training data, 2) the model type, and 3) the common evaluation metrics, including their inter-correlations and the nuances of LLM-as-a-judge approaches.

In conclusion, this work underscores the intricate relationship between MDS model architectures, the stylistic properties of data domains, and the evaluation metrics used to gauge performance. For models, a reduced stylistic bias and the ability to handle extensive context are beneficial for domain transfer. For evaluation, our findings strongly advocate for a multi-faceted approach, exercising caution with lexical metrics for advanced models, and a keen awareness of the specific definitions, potential biases, scaling limitations, and inter-correlation patterns of even sophisticated trained evaluators. Future research should continue to develop more robust and reliable evaluation strategies that are sensitive to the nuances of abstractive summarization across diverse domains and model capabilities.

\section*{Limitations}
\label{sec:limits}
Despite the rise in popularity of large, pre-trained decoder models, most fine-tuned multi-document summarization models are on the scale of a few hundred million parameters, e.g., BART-Large. 
We did not analyze the effect of model size or scale concerning summarization performance or domain transfer abilities. 

Another limitation is our focus on English, as all of the datasets and learned metrics (i.e., UniEval), were in English or trained on English data.

\section*{Ethical Considerations}
\label{sec:ethics}
We did not evaluate the model output and training datasets for \textit{absolute} factuality (i.e., correctness with respect to the world) or harmful language. 
We leave these tasks to future work or the incorporation of an additional step for model safety.

AI Assistants were used in the preparation of this work, specifically for paper revision, table formatting, and figure editing.

\section*{Acknowledgements}

\bibliography{custom, references, anthology}

\clearpage
\appendix

\section{Dataset Details}
\label{app:dataset_details}
We include three MDS datasets: Multi-News+ \citep{choi-etal-2024-multi-news}, Multi-XScience \citep{lu-etal-2020-multi-xscience}), and ConvoSumm-Reddit \citep{fabbri-etal-2021-convosumm}.
We keep the original train, validation, and test splits as the original papers for all datasets.
See \Cref{fig:convosumm_example,fig:multinews_example,fig:multixscience_example} for an example from each dataset and \Cref{tab:dataset-stats} for summary statistics.

\paragraph{Multi-News+.} This dataset has 37,057/4,576/4,648 examples for train, validation, and test sets, respectively.
We keep the \texttt{NEWLINE\_CHAR} symbol in the documents.

\paragraph{Multi-XScience.} The Multi-XScience dataset had many examples with missing source documents. 
In addition to skipping those with no source documents, we also removed non ``multi'' examples which had less than two source documents, resulting in 11523/1937/1901 examples for train, validation, and test sets, respectively. 
We skipped 18846/3129/3192 examples from train/val/test sets with less than two source documents (i.e., abstracts) and 20023/3383/3403 examples from train/val/test with completely missing source documents. 
We also replace all document-specific citations \texttt{\@cite\_5} with a more generalizable \texttt{\@cite}.

\paragraph{ConvoSumm.} While ConvoSumm features multiple conversation datasets, we focus on Reddit.
Similar to the original work, we flatten comment threads as individual source documents, resulting in 201/50/250 examples for train, validation, and test sets, respectively. 
We keep the \texttt{|NEWLINE|} symbols in the documents.

\section{Training Details}
\label{app:train_details}
Unless specified otherwise, each model was trained on a single Nvidia A100 GPU with 80GB of memory.
The training hyperparameters for PRIMERA, Summ$^N$, and ConvoSumm-arg-filtered are in \Cref{tab:train_primera,tab:train_summn,tab:train_convosumm}, respectively.
Below are the prompts and details for running Llama for the zero-shot and in-context 1-shot comparisons.

\paragraph{PRIMERA}
The authors' released models were from HuggingFace fine-tuned on Multi-XScience (\url{https://huggingface.co/allenai/PRIMERA-multixscience}) and MultiNews (\url{https://huggingface.co/allenai/PRIMERA-multinews}).
The authors' code is on GitHub (\url{https://github.com/allenai/PRIMER}).

\paragraph{Llama 3.1 8B-Instruct.}
Due to the large size of Llama 8B, we hosted it as a service with vLLM on 4 Nvidia A100 80GB GPUs \citep{kwon_efficient_2023}.\footnote{\url{https://docs.vllm.ai}}
We followed model examples on how to interact for the summarization task (\Cref{tab:llama_instructions}).

\section{Evaluation Details}
\label{app:evaluation}

\paragraph{Multi-dimensional analyses with Prometheus.}
We developed rubrics for Prometheus based on definitions set forth by prior work. We altered the definitions slightly for the multi-document use case. For all \textit{criteria}, i.e., description for model feedback, we use the same score rubric except for Factuality and Coverage:

\begin{enumerate}[align=left,label=Score \arabic*:,noitemsep]
    \item Very poor
    \item Poor
    \item Barely acceptable
    \item Good
    \item Very Good
\end{enumerate}

The prompt format is in \Cref{fig:prometheus_prompt} and the rubrics are in \Cref{fig:prometheus_criteria}.

The rubric for Coverage, inspired by \citet{wolhandler_how_2022}:
\begin{enumerate}[align=left,label=Score \arabic*:,noitemsep]
    \item The summary misses most salient information and/or fails to draw from many or all of the source documents.
    \item The summary includes some salient information, but frequently omits key points or fails to adequately represent all source documents.
    \item The summary covers a good portion of the salient information and represents most source documents, though some important details or content from some documents may be missing or underrepresented.
    \item The summary incorporates most salient information from all source documents, with only minor omissions of less critical details, and adequately represents all documents.
    \item The summary comprehensively and accurately captures all salient information from all source documents, providing a well-rounded representation that integrates content from all sources effectively.
\end{enumerate}

The rubric for Factuality, taken from the Prometheus code repository:
\begin{enumerate}[align=left,label=Score \arabic*:,noitemsep]
    \item The model's responses are mostly incorrect or based on unfounded information.
    \item The model sometimes provides factually correct responses, but inaccuracies are common.
    \item The model generally provides factually correct information, though some errors occur.
    \item The model often provides factually accurate information with only occasional minor errors.
    \item The model consistently provides responses that are factually correct and well-supported by evidence.
\end{enumerate}

\section{MDS Domain Generalizability}
\label{app:results-domain}

We compare the results from training and testing on a single domain to training on a domain and testing on the others (``From''), and training on other domains and testing on the domain (``To''). 
``Ref'' refers to the results when running the metrics on the ground-truth data. See \Cref{tab:results-data}.

\paragraph{Models trained on ConvoSumm-Reddit transfer well to other domains.} On average, models trained on ConvoSumm-Reddit and tested on the News and Science datasets see minimal changes in performance across quality and ground-truth similarity metrics. The exception to this trend is the large decrease in MoverScore, which is more sensitive to a performance drop of a hundredth of a decimal due to its often low scores (e.g., $0.05$).\footnote{We discuss MoverScore further in \Cref{sec:analysis-metrics}.}

\paragraph{Transferring from Multi-News+ leads to a decrease in factuality.} Compared to transferring from the other datasets, out-of-domain summaries generated with a model trained on Multi-News+ contain more factual errors (8-11\% more). This is most likely due to the large style bias of smaller models, which tend to hallucinate quotes (see \Cref{fig:sumn_summaries} for an example). This is different from the findings of \citep{belem-etal-2025-single}, who found that models (LLMs) hallucinate more on conversation data than news data.

\paragraph{Most metrics are impacted by models' bias to the style of the training dataset.} The style of source documents and summaries differ greatly in each dataset. The ConvoSumm-Reddit example from \Cref{fig:convosumm_example,fig:multinews_example} shows the stark difference in style between the types of summaries in ConvoSumm-Reddit and those from Multi-News+.
This difference in style explains the performance loss from Conversation and News domains, as the summary wording would be completely different and penalized by the ground-truth similarity metrics ($-31$ to $-9\%$ and $-92$ to $+2\%$, respectively). We confirm prior work's results that smaller generative models have more style bias to their training data, as shown by the large changes in ground-truth similarity and compression ratios.\footnote{Compression is a style metric because some datasets have more condensed summaries than others.}

\section{Raw Results}
\label{app:results-raw}

The domain-transfer results in \Cref{tab:results-model} show the summarized relative performance of the models.
The detailed performance across domains and models are in \Cref{tab:results-detailed}.

\input{dataset_examples}

\input{dataset_stats}
\input{correlation_figures}
\input{results_data}
\input{detailed_results}
\input{prometheus_prompts}
\input{unieval_criteria}
\input{llama_instructions}
\input{model_training_settings}
\input{transfer_examples}

\end{document}

%% file: fig1.tex
\newtcolorbox{promptboxhalf}[1]{
    enhanced,
    width=\columnwidth, 
    arc=1mm,
    boxrule=0.8pt,
    colback=white,
    colframe=boxframecolor,
    colbacktitle=boxtitlecolor,
    coltitle=black,
    fonttitle=\bfseries\sffamily,
    title=#1,
    before upper={\plexmono},
    boxsep=0.5pt,
    toptitle=1mm,
    bottomtitle=0mm,
}

\begin{figure}[]
    \small
    \centering
    \begin{promptboxhalf}{Domain-Transfer Example}
        \begin{tabularx}{\linewidth}{@{} >{\bfseries}p{1.7cm} >{\RaggedRight}X @{}}
        Ground-Truth & A couple of the responses to this post are either satirical or sarcastic. ... \\ 
        \midrule
        News-based &
        Might you be a little nervous about playing a video game for free? you're not alone. ... \\
        \midrule
        Science-based &
        @cite @cite have both mentioned that the reason for the lack of a sub is due to Microsoft's refusal to allow sub-free games on Xbox. ...
        \end{tabularx}
    \end{promptboxhalf}
    \caption{Excerpt generations two types of multi-document summarization models (Summ$^N$ and ConvoSumm), trained on different datasets (Multi-News+ and Multi-XScience), and both evaluated on a conversation summarization task (ConvoSumm-Reddit). The ROUGE-LSum of the generations to the Ground-Truth are 0.13 and 0.12, respectively, and UniEval-Relevancy scores of 0.78 and 0.3, respectively.}
    \label{fig:transfer_ex}
\end{figure}

%% file: dataset_examples.tex

\begin{figure*}[]
    \centering
    \begin{promptbox}{Example from Multi-News+}
        \begin{tabularx}{\linewidth}{@{} >{\bfseries\RaggedRight}p{2cm} >{\RaggedRight}X @{}}
        Source\hspace{0.5cm}Documents & 
        \begin{tabular}{@{}p{\linewidth}@{}}
        Washington ( ap ) — as a crucial second sign-up season gears up, the obama administration said sunday that healthcare.gov is stable and working well, a far cry from last year's frozen computer screens and frustrated customers. Health and human services secretary sylvia burwell smiles while answering a reporters question following her tour of the greater prince william community health center, evergreen terrace site, in manassas,... ( associated press ) health and human services secretary sylvia burwell gestures while answering a reporters question following her tour of the greater prince william community health center, evergreen terrace site, in manassas,... ( associated press ) health and human services secretary sylvia m. Burwell said she expects " strong and healthy growth " for 2015. About 7 million people are signed up, and burwell expects to grow that by 2 million more or so. The congressional budget office has projected a total of 13 million enrolled for 2015, and some see the administration as trying to lower expectations. Burwell told nbc's " meet the press " that 100,000 people had submitted new applications this weekend via the federal website serving 37 states. That's a big difference from last year, when only a handful of customers managed to enroll on the first day. Burwell also said that a half-million people who already have coverage through the program were able to log into their accounts this time. There were reports saturday that returning customers had problems, but some of that may have been confusion trying to remember user names and passwords. Administration spokesman aaron albright said sunday he had not seen any indication of problems. Burwell said call centers have taken 100,000 calls, another indication of consumer interest. President barack obama noted the improvements. " healthcare.gov works really well now, " he said. Healthcare.gov is an online marketplace that offers subsidized private coverage to people who don't have health insurance on the job. Because of political opposition and \\
        \cmidrule(r){1-1}
        Getty burwell: 100,000 new obamacare applications secretary of health and human services sylvia mathews burwell said on sunday 100,000 people submitted new applications for obamacare in the first days of the second open enrollment period. Speaking on nbc's "meet the press, " burwell offered an update to the numbers provided saturday when she said 23,000 applied in the first eight hours of the new open enrollment period. Story continued below in addition to new applicants, 500,000 customers were able to log on to healthcare.gov, and 1 million people have gone "window shopping " to compare insurance plans and prices over the past week. The numbers offer a stark contrast to the first enrollment period a year ago when healthcare.gov launched with severe technical problems that prevented people from applying. Authors:
        \end{tabular} \\
        \midrule
        Summary & After last year's dismal debut healthcare.gov is feeling much better now, and is busily connecting americans with health care plans as the second sign-up period opened yesterday, health and human services secretary sylvia burwell said today. The site took in 100,000 new applications in its first 24 hours, reports politico, saw 500,000 log in to their accounts, and 1 million more went " window-shopping " for plans. Call centers took 100,000 calls. Burwell said she expects " strong and healthy growth " this year — about another 2 million people to add to the 7 million already signed up, per the ap. A white house spokesman said today there was no indication of problems. " healthcare.gov works really well now, " said president obama. \\
        \end{tabularx}
    \end{promptbox}
    \caption{Example from the Multi-News+ dataset.}
    \label{fig:multinews_example}
\end{figure*}

\begin{figure*}[]
    \centering
    \begin{promptbox}{Example from Multi-XScience}
        \begin{tabularx}{\linewidth}{@{} >{\bfseries}l >{\RaggedRight}X @{}}
        Source Documents & 
        \begin{tabular}{@{}p{\linewidth}@{}}
        Despite the huge amount of recent research efforts on entity resolution (matching) there has not yet been a comparative evaluation on the relative effectiveness and efficiency of alternate approaches. We therefore present such an evaluation of existing implementations on challenging real-world match tasks. We consider approaches both with and without using machine learning to find suitable parameterization and combination of similarity functions. \dots \\
        \cmidrule(r){1-1}
        This tutorial provides a comprehensive and cohesive overview of the key research results in the area of record linkage methodologies and algorithms for identifying approximate duplicate records, and available tools for this purpose. It encompasses techniques introduced in several communities including databases, information retrieval, statistics and machine learning. It aims to identify similarities and differences across the techniques as well as their merits and limitations.
        \end{tabular} \\
        \midrule
        Summary & For accurate entity linking and deduplication, early works on the subject examined many methods such as cosine similarity match, distance based match, TF IDF, and Soundex. The well known similarity measures for entity linking are summarized and reviewed in \\cite{key}; and the work in \\cite{key} presents a comparative evaluation of some existing works. \\
        \end{tabularx}
    \end{promptbox}
    \caption{Example from the Multi-XScience dataset.}
    \label{fig:multixscience_example}
\end{figure*}
    
\begin{figure*}[]
    \centering
    \begin{promptbox}{Example from ConvoSumm-Reddit}
        \begin{tabularx}{\linewidth}{@{} >{\bfseries}l >{\RaggedRight}X @{}}
        Source Documents & 
        \begin{tabular}{@{}p{\linewidth}@{}}
        Title: Hand from tonight. Subreddit: poker Playing at my local pub league. Villian is loose aggressive. Sorry if this is bad formatting just quickly whipped it up for your opinions. \par Button: (~30000) \par SB: Villian: (50000) \dots \\
        \cmidrule(r){1-1}
        Shove pre but I imagine this is the sort of game where you get told off for doing that
        \end{tabular} \\
        \midrule
        Summary & A lot of commenters debate the semantics and word choice of another comment. Some commenters say that the poster should shove even though other players will not like that move very much. Other commenters discuss the differences of some poker terminology. \\
        \end{tabularx}
    \end{promptbox}
    \caption{Example from the ConvoSumm-Reddit dataset.}
    \label{fig:convosumm_example}
\end{figure*}

%% file: dataset_stats.tex
\begin{table*}[]
    \centering
    \resizebox{\textwidth}{!}{\begin{tabular}{@{}l l rrrr@{}} 
    \toprule
     &  & \multicolumn{1}{c}{\# Examples} & \multicolumn{1}{c}{\# Source Docs.} & \multicolumn{1}{c}{Summary Length} & \multicolumn{1}{c}{Avg. Doc. Length} \\ 
    \midrule 
     \multirow{3}{*}{ConvoSumm (Reddit)} & Test & 250 & 14.42 (5.55) & 57.78 (12.04) & 53.02 (26.60) \\
     & Train & 201 & 14.98 (6.28) & 58.32 (12.95) & 49.46 (26.38) \\
     & Val & 50 & 16.74 (7.21) & 59.18 (14.99) & 52.88 (27.29) \\ 
    \midrule 
    \multirow{3}{*}{MultiNews+} & Test & 4648 & 2.53 (0.88) & 230.94 (69.57) & 226.07 (64.15) \\
     & Train & 37057 & 2.55 (0.91) & 233.54 (69.98) & 224.30 (64.96) \\
     & Val & 4576 & 2.51 (0.86) & 231.45 (69.64) & 226.59 (63.35) \\
    \midrule
    \multirow{3}{*}{MultiXScience} & Test & 1901 & 3.55 (1.95) & 106.35 (41.60) & 195.32 (48.01) \\
     & Train & 11523 & 3.58 (2.05) & 106.82 (42.14) & 197.03 (47.91) \\
     & Val & 1937 & 3.55 (1.99) & 104.03 (41.38) & 196.98 (48.84) \\ 
    \bottomrule
    \end{tabular}}
    \caption{Statistics for each dataset used in this work: ConvoSumm-Reddit, Multi-News+, and Multi-XScience. The number of tokens in the (ground truth) summary and documents are measured with the BART tokenizer. Since we remove examples with missing information, our numbers might not match those in the original dataset papers (\Cref{app:dataset_details}).}
    \label{tab:dataset-stats}
\end{table*}

%% file: correlation_figures.tex
\begin{figure*}[htbp] 
    \centering 

    \begin{subfigure}{0.48\textwidth} 
        \includegraphics[width=\linewidth]{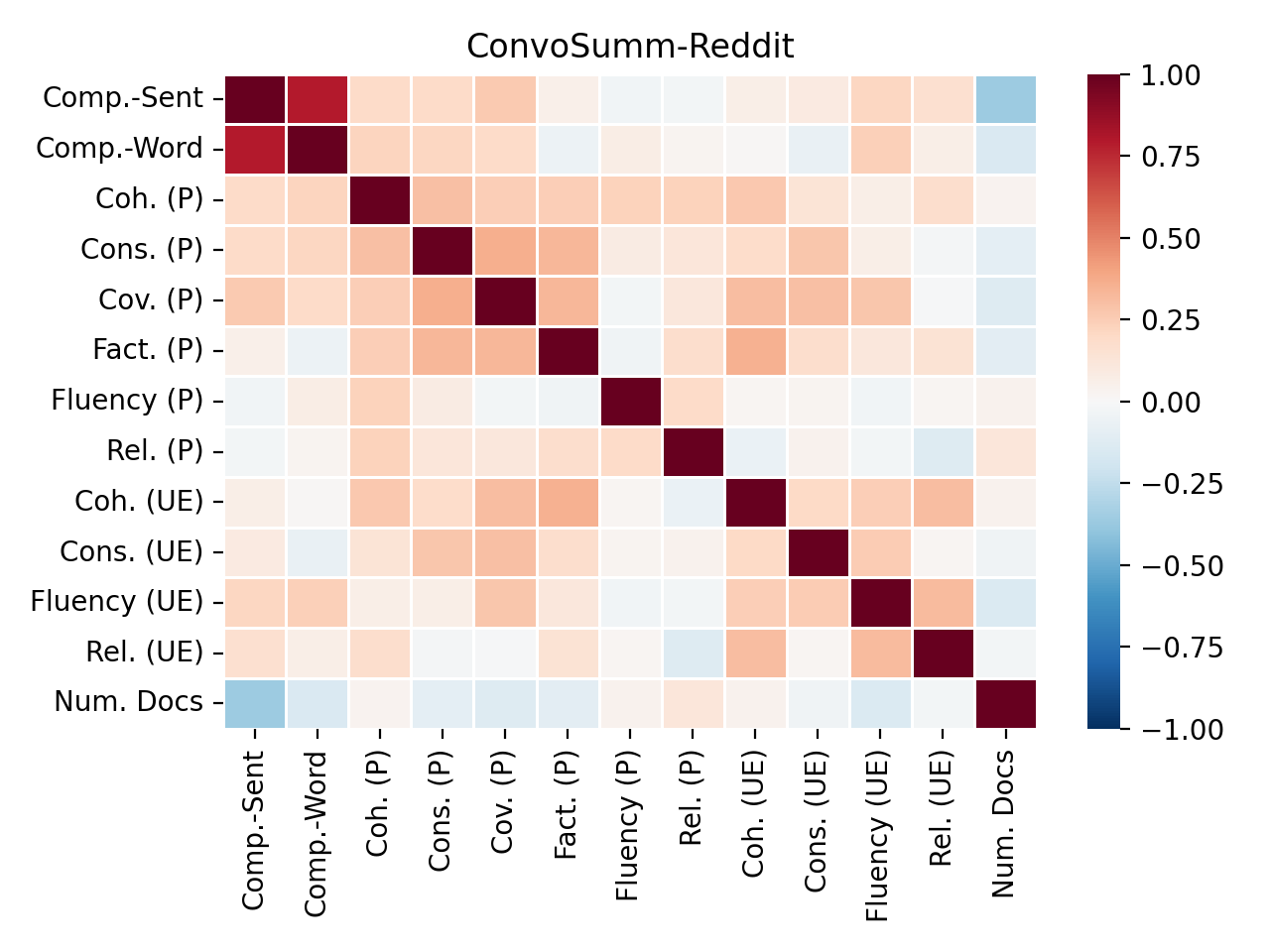}
        \caption{ConvoSumm-Reddit}
        \label{fig:corr-convo}
    \end{subfigure}
    \hfill 
    \begin{subfigure}{0.48\textwidth} 
        \includegraphics[width=\linewidth]{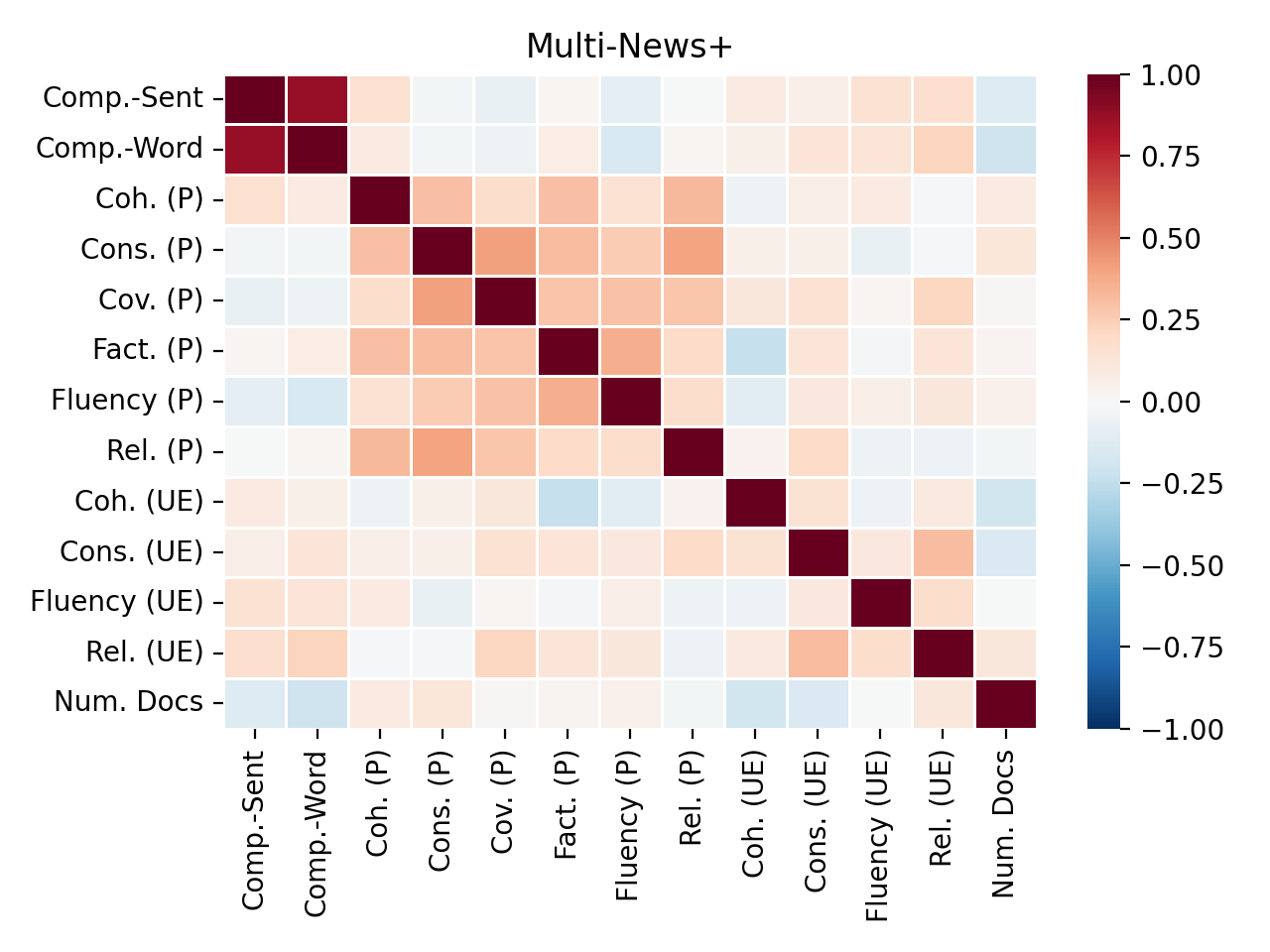}
        \caption{Multi-News+}
        \label{fig:corr-news}
    \end{subfigure}

    \vspace{1em} 

    \begin{subfigure}{0.48\textwidth} 
        \includegraphics[width=\linewidth]{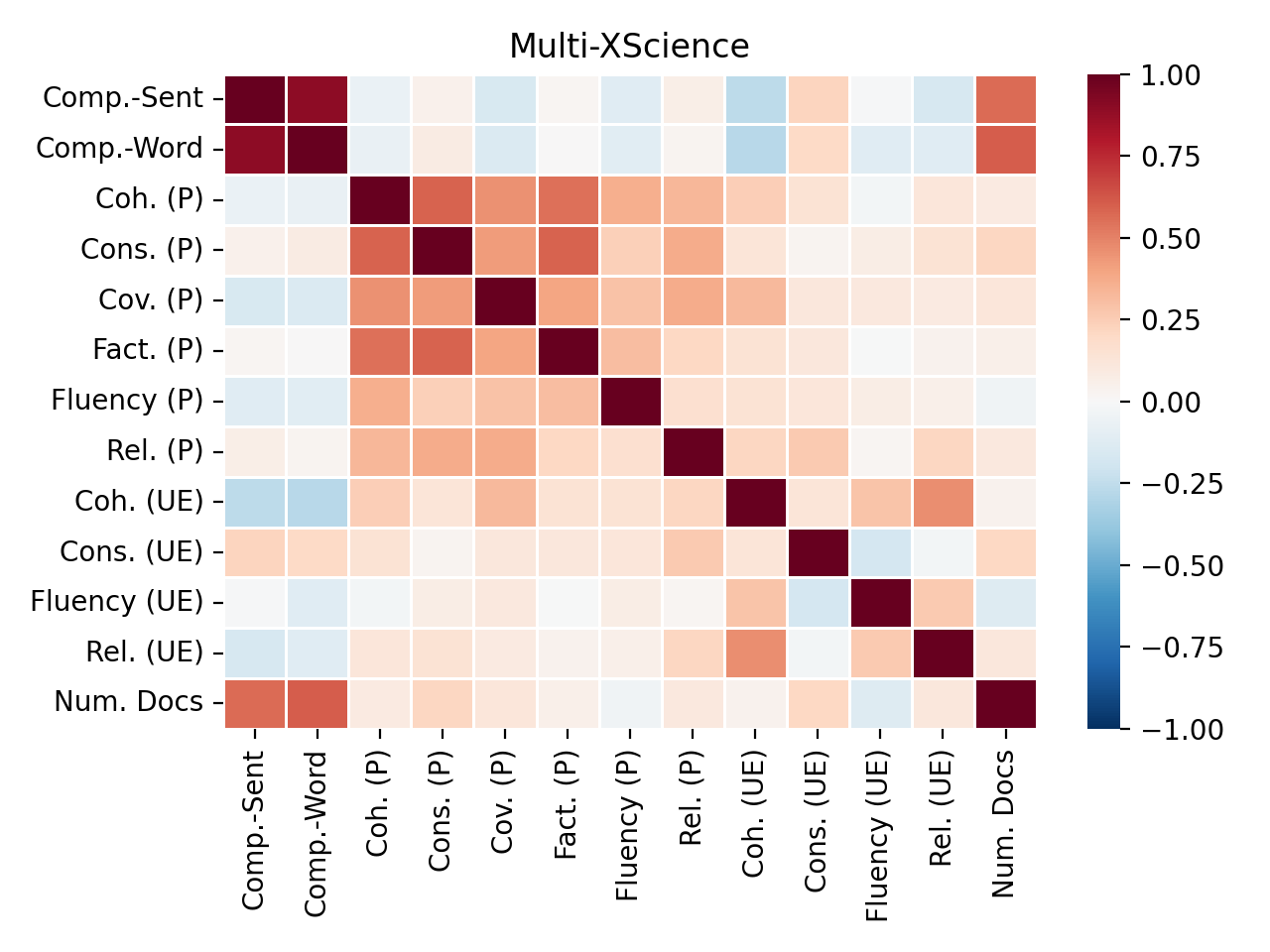} 
        \caption{Multi-XScience}
        \label{fig:corr-science}
    \end{subfigure}

    \caption{Pearson correlation between metrics on the test ground-truth (reference) for the evaluated datasets.}
    \label{fig:corr}
\end{figure*}

\begin{figure*}[htbp] 
    \centering 

    \begin{subfigure}{0.48\textwidth} 
        \includegraphics[width=\linewidth]{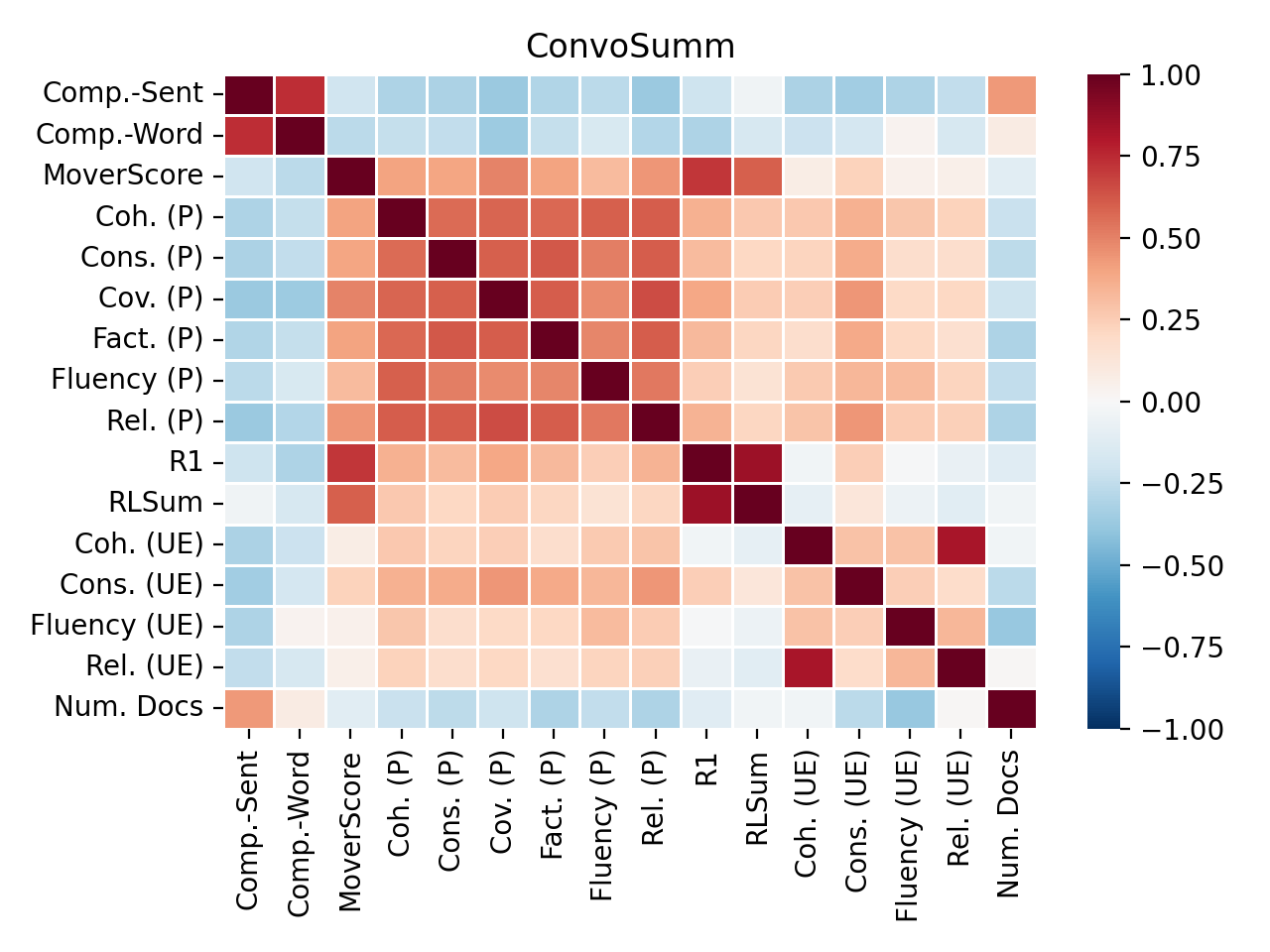}
        \caption{ConvoSumm}
        \label{fig:corr-convosumm}
    \end{subfigure}
    \hfill 
    \begin{subfigure}{0.48\textwidth} 
        \includegraphics[width=\linewidth]{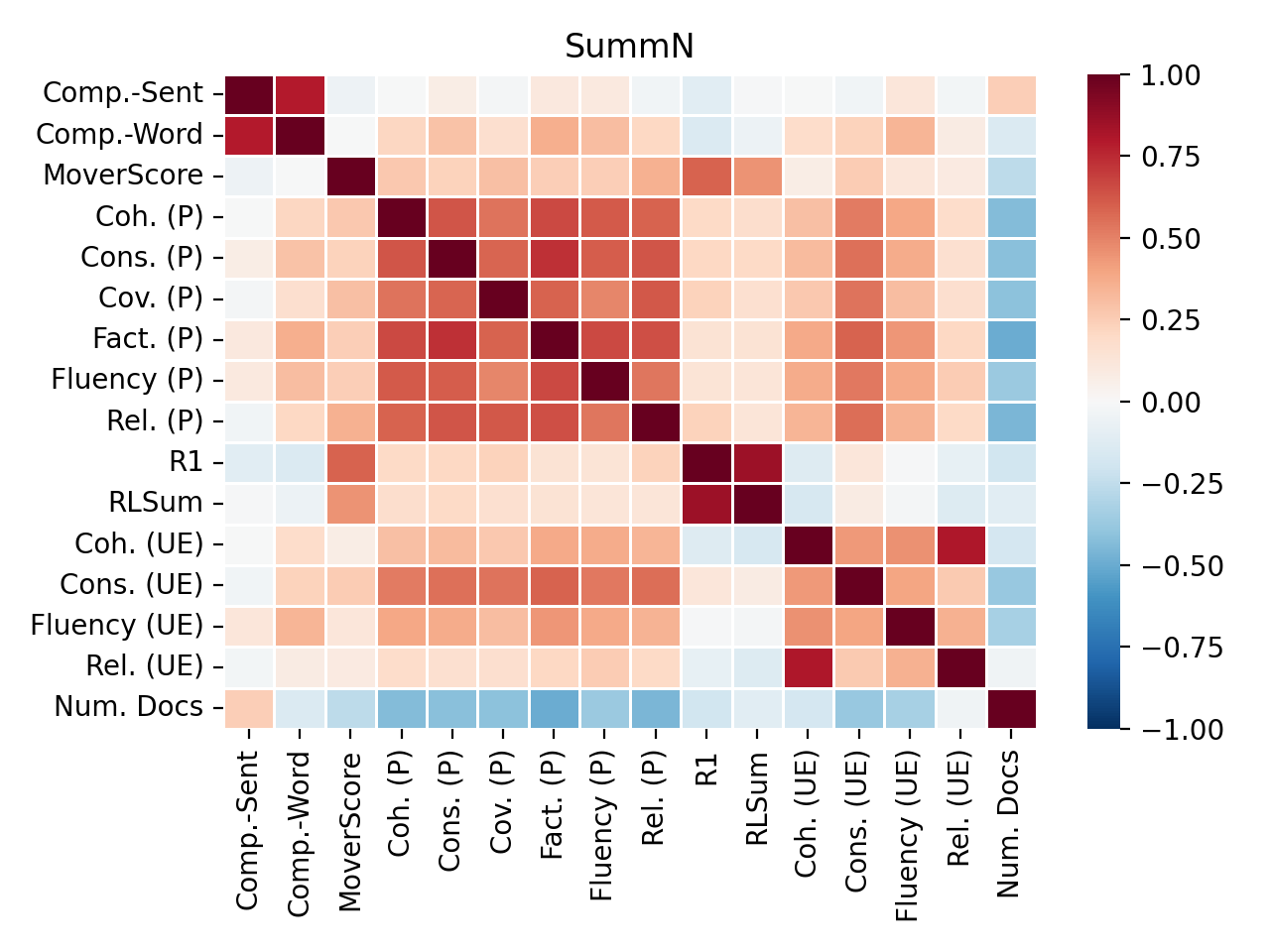}
        \caption{Summ$^{N}$}
        \label{fig:corr-summn}
    \end{subfigure}

    \vspace{1em} 

    \begin{subfigure}{0.48\textwidth} 
        \includegraphics[width=\linewidth]{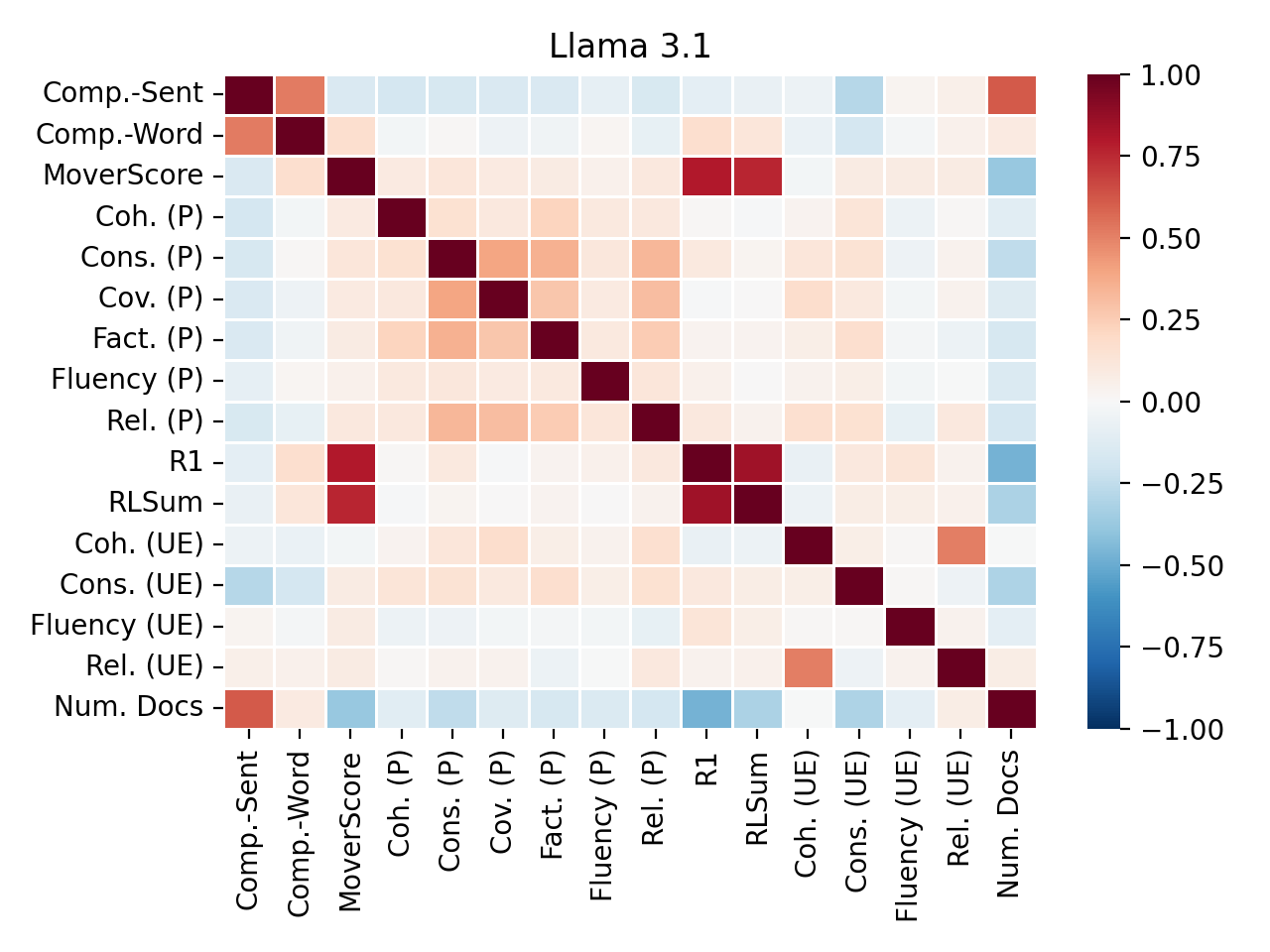} 
        \caption{Llama 3.1 8B}
        \label{fig:corr-llama}
    \end{subfigure}
    \hfill
    \begin{subfigure}{0.48\textwidth} 
        \includegraphics[width=\linewidth]{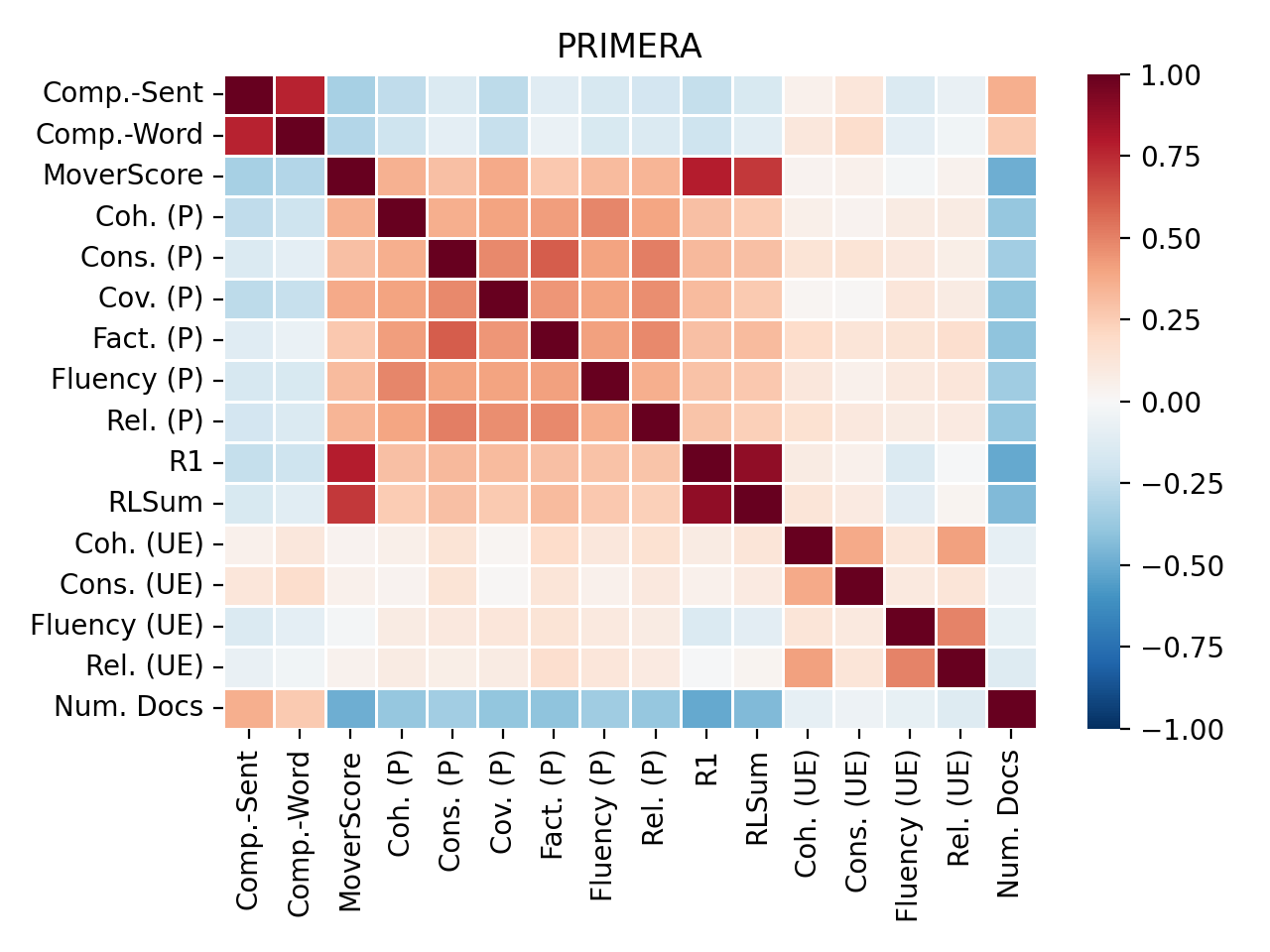} 
        \caption{PRIMERA}
        \label{fig:corr-primera}
    \end{subfigure}

    \caption{Pearson correlation between metrics across all datasets for each model in the domain-transfer setting.}
    \label{fig:corr_model}
\end{figure*}

%% file: results_data.tex
\begin{table*}[]
\resizebox{\textwidth}{!}{%
\begin{tabular}{@{}p{0.15\textwidth} l rrrr|rrrr|rrrr@{}}
\toprule
& & \multicolumn{4}{c}{ConvoSumm-Reddit} & \multicolumn{4}{c}{Multi-News+} & \multicolumn{4}{c}{Multi-XScience} \\
\cmidrule(lr){3-6} \cmidrule(lr){7-10} \cmidrule(lr){11-14}
&  & \multicolumn{1}{c}{} & \multicolumn{1}{c}{From $\Delta$(\%)} & \multicolumn{1}{c}{To $\Delta$(\%)} & \multicolumn{1}{c}{Ref.} & \multicolumn{1}{c}{} & \multicolumn{1}{c}{From $\Delta$(\%)} & \multicolumn{1}{c}{To $\Delta$(\%)} & \multicolumn{1}{c}{Ref.} & \multicolumn{1}{c}{} & \multicolumn{1}{c}{From $\Delta$(\%)} & \multicolumn{1}{c}{To $\Delta$(\%)} & \multicolumn{1}{c}{Ref.} \\
\midrule
\multirow{3}{=}{Factuality} & Cons. (UE) & 0.74 & \textbf{+11\%} & -9\% & 0.71 & 0.80 & -5\% & +8\% & 0.70 & 0.72 & -5\% & +1\% & 0.55 \\
& Fact. (P)    & 2.25 & \textbf{+11\%} & -21\% & 3.07 & 2.46 & -20\% & +10\% & 2.84 & 2.11 & -5\% & -1\% & 1.82 \\ 
& Cons. (P)   & 2.42 & \textbf{+5\%} & -18\% & 3.06 & 2.62 & -21\% & +11\% & 3.16 & 2.09 & +1\% & -7\% & 1.86 \\ 
\midrule
\multirow{4}{=}{Ground-Truth Similarity} & Rel. (UE) & 0.69 & -1\% & -15\% & 0.99 & 0.73 & 0\% & -17\% & 0.98 & 0.63 & -29\% & +5\% & 0.99 \\
& R1          & 0.26 & \textbf{+13\%} & -30\% & -1.00 & 0.42 & -46\% & +28\% & -1.00 & 0.26 & +6\% & -21\% & -1.00 \\
& RLSum       & 0.17 & -1\% & -22\% & -1.00 & 0.21 & -37\% & +11\% & -1.00 & 0.15 & \textbf{+3\%} & -20\% & -1.00 \\
& MoverScore          & 0.05 & -238\% & -98\% & 1.00 & 0.17 & -80\% & -110\% & 1.00 & 0.05 & -36\% & -107\% & 1.00 \\
\midrule
\multirow{5}{=}{Quality} & Rel. (P)    & 2.62 & \textbf{+5\%} & -18\% & 3.23 & 2.81 & -15\% & +8\% & 3.39 & 2.37 & -4\% & -2\% & 2.78 \\ 
& Fluency (UE) & 0.85 & +2\% & -5\% & 0.90 & 0.76 & \textbf{+4\%} & -4\% & 0.76 & 0.77 & -9\% & +7\% & 0.66 \\
& Fluency (P)   & 2.69 & \textbf{+7\%} & -16\% & 3.30 & 2.79 & -21\% & +11\% & 2.75 & 2.17 & +2\% & -5\% & 1.53 \\ 
& Coh. (UE) & 0.73 & \textbf{+6\%} & -7\% & 0.73 & 0.81 & -3\% & -3\% & 0.75 & 0.62 & -13\% & +2\% & 0.43 \\
& Coh. (P)    & 2.60 & \textbf{+6\%} & -19\% & 3.39 & 2.89 & -24\% & +12\% & 3.70 & 2.17 & +2\% & -7\% & 1.51 \\ 
& Comp.-Sent    & 9.01 & \textbf{+45\%} & -72\% & 12.03 & 2.74 & -44\% & +19\% & 2.08 & 4.07 & -42\% & +18\% & 6.10 \\
& Comp.-Word    & 7.18 & \textbf{+13\%} & -24\% & 8.63 & 2.47 & -18\% & +8\% & 1.98 & 4.40 & -8\% & +4\% & 5.92 \\
\bottomrule
\end{tabular}}
\caption{Multi-document summarization (MDS) model performance and relative performance \textbf{aggregated by domain} when trained and tested on the same domain, and the relative change ($\Delta$) when transferred to other domains (From), and vice-versa (To). A relative change of over $1$ is possible with MoverScore because of its $[-1,1]$ range. For all metrics other than Compression, a higher score is better. Metrics measured by UniEval and Prometheus are marked by (UE) and (P), respectively.}
    \label{tab:results-data}
\end{table*}

%% file: detailed_results.tex
\begin{table*}[]
\centering
\resizebox{\textwidth}{!}{\begin{tabular}{@{}lll|rrr|rrrr|rrrrrrr|rr@{}} 
\toprule
\multicolumn{1}{c}{} & \multicolumn{2}{c}{Domain} & \multicolumn{3}{c}{Factuality} & \multicolumn{4}{c}{Ground-Truth Similarity} & \multicolumn{8}{c}{Quality} \\
\cmidrule(lr){4-6} \cmidrule(lr){7-10} \cmidrule(lr){11-18}
& \multicolumn{1}{c}{Train} & \multicolumn{1}{c}{Test} &
\multicolumn{1}{c}{C(UE)} & \multicolumn{1}{c}{F(P)} & \multicolumn{1}{c}{C(P)} &
\multicolumn{1}{c}{R(UE)} & \multicolumn{1}{c}{R1} & \multicolumn{1}{c}{RLS} & \multicolumn{1}{c}{MS} &
\multicolumn{1}{c}{R(P)} & \multicolumn{1}{c}{Fl(UE)} & \multicolumn{1}{c}{Fl(P)} & \multicolumn{1}{c}{Co(UE)} & \multicolumn{1}{c}{Co(P)} & \multicolumn{1}{c}{Cov(P)} &
\multicolumn{1}{c}{C.Sent} & \multicolumn{1}{c}{C.Word} \\
\midrule
\multirow{6}{*}{ConvoSumm} & Convo & News+ & 0.75 & 1.83 & 1.90 & 0.51 & 0.24 & 0.14 & 0.00 & 2.08 & 0.91 & 2.02 & 0.57 & 1.94 & 1.68 & -4.31 & -7.54 \\
& Convo & Science & 0.67 & 1.95 & 1.97 & 0.61 & 0.23 & 0.14 & 0.02 & 2.27 & 0.91 & 2.22 & 0.58 & 2.03 & 1.76 & -5.17 & -9.23 \\ \cmidrule{2-18}
& News+ & Convo & 0.68 & 1.59 & 1.78 & 0.75 & 0.23 & 0.13 & 0.01 & 1.97 & 0.70 & 1.72 & 0.78 & 1.74 & 1.77 & -3.62 & -2.51 \\
& News+ & Science & 0.69 & 1.77 & 1.76 & 0.70 & 0.23 & 0.13 & 0.04 & 2.13 & 0.81 & 1.79 & 0.71 & 1.87 & 1.89 & -2.31 & -2.32 \\ \cmidrule{2-18}
& Science & Convo & 0.44 & 1.22 & 1.26 & 0.26 & 0.22 & 0.15 & -0.01 & 1.34 & 0.60 & 1.27 & 0.26 & 1.29 & 1.25 & -9.32 & -7.78 \\
& Science & News+ & 0.66 & 1.81 & 1.87 & 0.26 & 0.30 & 0.16 & 0.04 & 2.10 & 0.64 & 1.65 & 0.30 & 1.70 & 1.69 & -3.37 & -3.99 \\
\midrule
\multirow{6}{*}{PRIMERA} & Convo & News+ & 0.83 & 1.97 & 2.08 & 0.41 & 0.38 & 0.18 & 0.13 & 2.38 & 0.63 & 2.04 & 0.71 & 1.94 & 2.01 & -1.99 & -1.28 \\
& Convo & Science & 0.84 & 1.96 & 1.95 & 0.61 & 0.23 & 0.12 & 0.06 & 2.38 & 0.79 & 1.82 & 0.73 & 1.79 & 1.94 & -2.23 & -1.59 \\ \cmidrule{2-18}
& News+ & Convo & 0.82 & 1.34 & 1.53 & 0.40 & 0.19 & 0.11 & -0.01 & 1.76 & 0.67 & 1.36 & 0.72 & 1.26 & 1.52 & -5.29 & -3.17 \\
& News+ & Science & 0.82 & 2.03 & 1.96 & 0.65 & 0.23 & 0.12 & 0.06 & 2.40 & 0.82 & 1.92 & 0.74 & 1.83 & 2.04 & -2.40 & -1.61 \\ \cmidrule{2-18}
& Science & Convo & 0.80 & 1.43 & 1.49 & 0.41 & 0.19 & 0.11 & -0.02 & 1.77 & 0.64 & 1.34 & 0.67 & 1.28 & 1.58 & -4.46 & -2.61 \\
& Science & News+ & 0.78 & 1.96 & 2.07 & 0.44 & 0.38 & 0.18 & 0.13 & 2.33 & 0.62 & 1.95 & 0.68 & 1.94 & 1.99 & -2.42 & -1.51 \\
\midrule
\multirow{6}{*}{Summ$^N$} & Convo & News+ & 0.90 & 2.71 & 2.64 & 0.64 & 0.23 & 0.14 & 0.01 & 2.75 & 0.89 & 2.92 & 0.82 & 2.74 & 2.18 & -5.47 & -8.28 \\
& Convo & Science & 0.77 & 2.51 & 2.36 & 0.72 & 0.21 & 0.13 & 0.01 & 2.48 & 0.92 & 2.61 & 0.88 & 2.27 & 1.99 & -7.72 & -11.28 \\ \cmidrule{2-18}
& News+ & Convo & 0.67 & 1.44 & 1.58 & 0.71 & 0.22 & 0.13 & 0.00 & 1.95 & 0.70 & 1.74 & 0.73 & 1.58 & 1.67 & -3.45 & -2.16 \\
& News+ & Science & 0.64 & 1.57 & 1.54 & 0.67 & 0.21 & 0.12 & 0.03 & 2.15 & 0.77 & 1.63 & 0.68 & 1.68 & 1.83 & -1.39 & -1.26 \\ \cmidrule{2-18}
& Science & Convo & 0.39 & 1.13 & 1.21 & 0.33 & 0.21 & 0.14 & -0.04 & 1.26 & 0.63 & 1.37 & 0.33 & 1.25 & 1.26 & -6.52 & -4.14 \\
& Science & News+ & 0.65 & 1.95 & 2.04 & 0.33 & 0.31 & 0.17 & 0.04 & 2.36 & 0.69 & 2.05 & 0.35 & 2.07 & 1.90 & -3.14 & -3.58 \\
\midrule
\multirow{6}{*}{Llama 3.1} & Convo & News+ & 0.87 & 3.24 & 3.56 & 0.92 & 0.40 & 0.21 & 0.16 & 3.64 & 0.94 & 4.11 & 0.95 & 3.98 & 3.02 & -2.31 & -1.79 \\
& Convo & Science & 0.88 & 3.32 & 3.47 & 0.91 & 0.24 & 0.15 & 0.08 & 3.58 & 0.94 & 4.06 & 0.97 & 4.00 & 3.23 & -1.39 & -1.29 \\ \cmidrule{2-18}
& News+ & Convo & 0.82 & 3.01 & 3.23 & 0.94 & 0.22 & 0.15 & 0.05 & 3.43 & 0.94 & 3.99 & 0.96 & 3.92 & 2.96 & -3.88 & -1.36 \\
& News+ & Science & 0.88 & 3.38 & 3.45 & 0.92 & 0.24 & 0.15 & 0.09 & 3.61 & 0.93 & 4.03 & 0.98 & 4.00 & 3.19 & -1.30 & -1.26 \\ \cmidrule{2-18}
& Science & Convo & 0.82 & 3.17 & 3.21 & 0.94 & 0.22 & 0.15 & 0.06 & 3.47 & 0.93 & 3.94 & 0.96 & 3.92 & 2.91 & -4.00 & -1.46 \\
& Science & News+ & 0.87 & 3.25 & 3.54 & 0.94 & 0.40 & 0.22 & 0.16 & 3.64 & 0.95 & 4.07 & 0.95 & 3.98 & 2.93 & -2.15 & -1.69 \\
\midrule
\multirow{3}{*}{Llama 3.1 (0-shot)} & zero-shot & Convo & 0.83 & 3.12 & 3.31 & 0.94 & 0.22 & 0.15 & 0.05 & 3.44 & 0.93 & 4.01 & 0.96 & 3.87 & 2.79 & -3.74 & -1.40 \\
& zero-shot & News+ & 0.88 & 3.26 & 3.41 & 0.93 & 0.41 & 0.22 & 0.16 & 3.57 & 0.94 & 4.04 & 0.94 & 4.00 & 3.02 & -2.07 & -1.64 \\
& zero-shot & Science & 0.87 & 3.42 & 3.52 & 0.93 & 0.24 & 0.16 & 0.08 & 3.47 & 0.93 & 4.06 & 0.98 & 3.96 & 3.29 & -1.38 & -1.30 \\
\bottomrule
\end{tabular}}
\caption{Multi-document summarization (MDS) model performance when trained and tested on the \textbf{different} domains. Results are on the test splits of each dataset. For all metrics, a greater score is better.}
\label{tab:results-detailed}
\end{table*}

%% file: prometheus_prompts.tex
\begin{figure*}[h!]
    \centering 

    \begin{promptbox}{Prometheus Evaluator Prompt}
        \begin{tabularx}{\linewidth}{@{} l >{\RaggedRight}X @{}}
        
        \bfseries System & 
        You are a fair judge assistant tasked with providing clear, objective feedback based on specific criteria, ensuring each assessment reflects the absolute standards set for performance. \\
        
        \midrule 
        
        \bfseries Assistant & 
        \textbf{\#\#\#Task Description:} \par
        An instruction (might include an Input inside it), a response to evaluate, a reference answer that gets a score of 5, and a score rubric representing a evaluation criteria are given.
        \begin{enumerate}[leftmargin=*, itemsep=1pt, topsep=3pt]
            \item Write a detailed feedback that assess the quality of the response strictly based on the given score rubric, not evaluating in general.
            \item After writing a feedback, write a score that is an integer between 1 and 5. You should refer to the score rubric.
            \item The output format should look as follows: "(write a feedback for criteria) [RESULT] (an integer number between 1 and 5)"
            \item Please do not generate any other opening, closing, and explanations.
        \end{enumerate}
        
        \textbf{\#\#\#The instruction to evaluate:} \par
        You are a multi-document summarizer. When given a list of DOCUMENTS, you provide a SUMMARY that incorporates all of the documents. Write a summary that incorporates all of the following documents: \par \medskip
        
        \textbf{\#\#\# DOCUMENTS} \par
        \texttt{\{all source documents\}} \par \medskip
        
        \textbf{\#\#\#Response to evaluate:} \par
        \texttt{\{model-generated summary\}} \par \medskip
        
        \textbf{\#\#\#Reference Answer (Score 5):} \par
        \texttt{\{ground-truth summary\}} \par \medskip
        
        \textbf{\#\#\#Score Rubrics:} \par
        \texttt{\{rubric\}} \par \medskip
        
        \textbf{\#\#\#Feedback:} \\
        
        \end{tabularx}
    \end{promptbox}
    
    \caption{The formatting is from \citet{kim-etal-2024-prometheus}, and the instructions and rubric are from \Cref{fig:prometheus_criteria}.}
    \label{fig:prometheus_prompt}
\end{figure*}

\begin{figure*}[]
    \centering
    \begin{promptbox}{Evaluation Criteria for Prometheus}
        \begin{tabularx}{\linewidth}{@{} >{\bfseries}l >{\RaggedRight}X @{}}
        Factuality & Is the model's summary factually correct and well-supported by the DOCUMENTS? \\ \midrule
        Relevance & How well does the summary capture the key points of the documents? Consider whether all and only the important aspects are contained in the summary. \textit{Only evaluate the summary for relevance. Do not penalize the summary for issues that are outside the scope of relevance.} \\ \midrule
        Consistency & How consistent are the facts in the summary with the facts in the original documents? Consider whether the summary does reproduce all facts accurately and does not make up untrue information. \textit{Do not use your own knowledge, and consider the documents as truth. Only evaluate the summary for consistency. Do not penalize the summary for issues that are outside the scope of consistency.} \\ \midrule
        Fluency & What is the quality of individual sentences, are they well-written and grammatically correct? Consider the quality of individual sentences.\textit{Only evaluate the summary for fluency. Do not penalize the summary for issues that are outside the scope of fluency.} \\ \midrule
        Coherence & What is the quality of all sentences collectively? How well do they fit together, and do they sound natural? Consider the quality of the summary as a whole. \textit{Only evaluate the summary for coherence. Do not penalize the summary for issues that are outside the scope of coherence.} \\ \midrule
        Coverage & Does the model's summary incorporate salient information from all the documents? \textit{Do not penalize the summary for issues that are outside the scope of coverage.}
        \end{tabularx}
    \end{promptbox}
    \caption{Evaluation dimensions provided to Prometheus \citep{kim-etal-2024-prometheus}. The criteria for Relevance, Consistency, Fluency, and Coherence are drawn from SummEval \citep{fabbri-etal-2021-summeval}. We added the \textit{italicized} text to the prompt to improve Prometheus performance.}
    \label{fig:prometheus_criteria}
\end{figure*}

%% file: unieval_criteria.tex
\begin{figure*}[]
    \centering
    \begin{promptbox}{Evaluation Criteria for UniEval (Summarization)}
        \begin{tabularx}{\linewidth}{@{} >{\bfseries}l >{\RaggedRight}X @{}}
        Relevance & Is this summary relevant to the reference? \\ \midrule
        Consistency & Is this claim consistent with the document? \\ \midrule
        Fluency & Is this a fluent paragraph? \\ \midrule
        Coherence & Is this a coherent summary to the document? \\
        \end{tabularx}
    \end{promptbox}
    \caption{Evaluation dimensions for UniEval \citep{zhong-etal-2022-towards}. Only Relevance requires access to a reference summary.}
    \label{fig:unieval_criteria}
\end{figure*}

%% file: llama_instructions.tex
\begin{table*}[]
    \centering
    \begin{promptbox}{Llama 3.1 Instruct Prompt Templates}
        \begin{tabularx}{\linewidth}{@{} >{\RaggedRight}X @{}}
        \bfseries Zero-shot \\
        \midrule
        \texttt{<|begin\_of\_text|><|start\_header\_id|>}user\texttt{<|end\_header\_id|>} \newline
        Write a summary that incorporates all of the following documents: \newline
        Document: \{source document 1\} \newline
        \dots \newline
        Document: \{source document $n$\} \newline
        \texttt{<|eot\_id|><|start\_header\_id|>}assistant\texttt{<|end\_header\_id|>} \\
        \midrule \midrule
        \bfseries In-context Learning \\
        \midrule
        \texttt{<|begin\_of\_text|><|start\_header\_id|>}user\texttt{<|end\_header\_id|>} \newline
        Write a summary that incorporates all of the following documents: \newline
        Document: \{source document 1\} \newline
        \dots \newline
        Document: \{source document $n$\} \newline
        \texttt{<|eot\_id|><|start\_header\_id|>}assistant\texttt{<|end\_header\_id|>} \newline
        \{target summary\} \newline
        \texttt{<|begin\_of\_text|><|start\_header\_id|>}user\texttt{<|end\_header\_id|>} \newline
        Write a summary that incorporates all of the following documents: \newline
        Document: \{source document 1\} \newline
        \dots \newline
        Document: \{source document $n$\} \newline
        \texttt{<|eot\_id|><|start\_header\_id|>}assistant\texttt{<|end\_header\_id|>}
        \end{tabularx}
    \end{promptbox}
    \caption{Prompt templates for Llama 3.1 8B Instruct for the zero-shot (top) and single-shot in-context learning tasks (bottom).}
    \label{tab:llama_instructions}
\end{table*}

%% file: model_training_settings.tex
\begin{table*}[]
    \centering
    \begin{tabular}{llrrr}
    \toprule
    Stage & Setting & Multi-News+ & ConvoSumm-Reddit & Multi-XScience \\
     \midrule
     General & \\
     & total number of updates & 30000 & 5000 & - \\
     & warmup updates & 500 & 200 & - \\
     & learning rate & 3e-5 & - & - \\
     & max input sequence & 1024 & - & - \\
     & max tokens (in batch) & 2048 & - & - \\
     & update frequency & 4 & - & - \\
     & patience & 2 & - & - \\
     & label smoothing & 0.1 & - & - \\
     & dropout & 0.1 & - & - \\
     & attention dropout & 0.1 & - & - \\
     & weight decay & 0.01 & - & - \\
     & optimizer & \fitcol{0.2}{Adam betas=(0.9, 0.999), eps=1e-8} & - & - \\
     & LR scheduler & polynomial decay & - & - \\
     & Gradient clipping & 0.1 & - & - \\     
     \midrule 
     Stage 1 & \\
        & max input sequence & 1024 & - & - \\
        & beam & 1 & - & - \\
        & length penalty & 1 & - & - \\
        & max length & 450 & - & - \\
        & min length & 0 & - & - \\
        & no-repeat-ngram-size & 3 & - & - \\
        & temperature & 0.9 & - & - \\
     \midrule 
     Stage 2 & \\
        & max input sequence & 1024 & - & - \\
        & beam & 1 & - & - \\
        & length penalty & 1 & - & - \\
        & max length & 550 & - & - \\
        & min length & 0 & - & - \\
        & no-repeat-ngram-size & 3 & - & - \\
        & temperature & 0.9 & - & - \\
     \midrule 
     Stage 3 & \\
        & max input sequence & 1024 & - & - \\
        & beam & 1 & - & - \\
        & length penalty & 1 & - & - \\
        & max length & 900 & 1024 & - \\
        & min length & 0 & - & - \\
        & no-repeat-ngram-size & 3 & - & - \\
        & temperature & 0.9 & - & - \\
    \bottomrule
    \end{tabular}
    \caption{Training details for Summ$^N$ on all datasets. Weights initialized from BART-Large-CNN. An entry of ``-'' indicates the same setting as the previous column.}
    \label{tab:train_summn}
\end{table*}

\begin{table*}[]
    \centering
    \begin{tabular}{llrrr}
    \toprule
    Setting & Multi-News+ & ConvoSumm-Reddit & Multi-XScience* \\
     \midrule
     max input sequence & 4096 & 4096 & 4096 \\
     total number of steps & 30K & 5K & 20K \\
     warmup steps & 1K & 200 & 2K \\
     optimizer & Adam & Adam & Adam \\
     learning rate & 3e-5 & 3e-5 & 3e-5 \\
     batch size & 16 & 16 & 16 \\
     seed & 0 & 42 & 0 \\
     label smoothing & 0.1 & 0.1 & 0.1 \\
     \midrule
     max length & 1024 & 1024 & 1024 \\
     min length & 0 & 0 & 0 \\
     beam & 1 & 1 & 1 \\
    \bottomrule
    \end{tabular}
    \caption{Training details for PRIMERA on all datasets. Models marked with (*) were trained by \citep{xiao-etal-2022-primera} and only available training information is filled in. Weights initialized from BART-Large.}
    \label{tab:train_primera}
\end{table*}

\begin{table*}[]
    \centering
    \begin{tabular}{llrrr}
    \toprule
    Setting & Multi-News+ & ConvoSumm-Reddit* & Multi-XScience \\
    \midrule
     max input sequence & 2048 & 2048 & 2048 \\
    max tokens (in batch) & 2048 & - & - \\
     update frequency & 1 & 1 & 1 \\
     total number of updates & 30000 & 200 & 200 \\
     warmup updates & 500 & 20 & 20 \\
     learning rate & 3e-5 & 3e-5 & 3e-5 \\
     label smoothing & 0.1 & 0.1 & 0.1 \\
     dropout & 0.1 & - & - \\
     attention dropout & 0.1 & - & - \\
     weight decay & 0.01 & - & - \\
     optimizer & \fitcol{0.2}{Adam betas=(0.9, 0.999), eps=1e-8} & - & - \\
     LR scheduler & polynomial decay & - & - \\
     Gradient clipping & 0.1 & - & - \\   
     \midrule
     max length & 1024 & 1024 & 1024 \\
     min length & 0 & 0 & 0 \\
     beam & 1 & 1 & 1 \\
     length penalty & 1 & 1 & 1 \\
    \bottomrule
    \end{tabular}
    \caption{Training details for ConvoSumm on all datasets. Models marked with (*) were trained by \citep{fabbri-etal-2021-convosumm} and only available training information is filled in. Weights initialized from BART-Large. An entry of ``-'' indicates the same setting as the previous column.}
    \label{tab:train_convosumm}
\end{table*}

%% file: transfer_examples.tex
\newcommand{\modelmetrics}[2]{%
    \textbf{#1} \par \medskip \RaggedRight #2%
}

\begin{figure*}
    \small
    \centering
    \begin{promptbox}{Generated Summaries: PRIMERA Models}
        \begin{tabularx}{\linewidth}{@{} >{\RaggedRight}p{2.9cm} >{\RaggedRight}X @{}}
        \modelmetrics{Multi-XScience}{Comp-Word: 3.06 \newline R-LSum: 0.16 \newline Fact. (P): 2.00 \newline Fact. (U): 0.80 \newline Fluency (P): 2.00 \newline Fluency (U): 0.53} & 
        Washington ( ap ) — as a crucial second sign-up season gears up, the obama administration said sunday that healthcare.gov is stable and working well, a far cry from last year's frozen computer screens and frustrated customers. Health and human services secretary sylvia burwell smiles while answering a reporters question following her tour of the greater prince william community health center, evergreen terrace site, in manassas,... ( associated press ) health and human service secretary syLVia burrell gestures while answeringa reporters question after her tour... ( associated Press ) health care secretary sylvia burwwind is an employee of the federal government who works to make healthcare.... (associated press ) Getty.Getty burwell: 100,000 new obamacare applications secretary of health and h... (Associated press ) healthcare.hgov is an online marketplace that offers subsidized private coverage to people who don't have health insurance on the job.Because of political opposition and the \\
        \midrule

        \modelmetrics{ConvoSumm}{Comp-Word: 2.04 \newline R-LSum: 0.18 \newline Fact. (P): 2.00 \newline Fact. (U): 0.91 \newline Fluency (P): 2.00 \newline Fluency (U): 0.67} &
        Washington ( ap ) — as a crucial second sign-up season gears up, the obama administration said sunday that healthcare.gov is stable and working well, a far cry from last year's frozen computer screens and frustrated customers. Health and human services secretary sylvia burwell smiles while answering a reporters question following her tour of the greater prince william community health center, evergreen terrace site, in manassas,... ( associated press ) health and human rights secretary sylivia burforth gestures while answering her tour in the greater Prince william Community health center during the first days of the second open enrollment period.Health and Human Services Secretary Sylvia Mathews Burwell also said a half-million people who already have coverage through the program were able to log into their accounts this time, compared to last year when thousands struggled to do so.Burwell said call centers have taken 100,000 calls, another indication of consumer interest.President barack obama noted the improvements.Healthcare.gov's numbers were down slightly from last season's first open enrollment.Officials said last year many people would not sign up because of the website problems.Aaron Albright, an Obama administration spokesman, said health and Human services' healthcare.org site " is accessible for anyone seeking to access the site" and the agency has been "redesigned and upgraded" for this year.Last week, the site's status improved significantly.Reuters contributed to this report.. \\
        \midrule

        \modelmetrics{MultiNews+}{Comp-Word: 2.80 \newline R-LSum: 0.19 \newline Fact. (P): 3.00 \newline Fact. (U): 0.83 \newline Fluency (P): 3.00 \newline Fluency (U): 0.74} &
        Washington ( ap ) — as a crucial second sign-up season gears up, the obama administration said sunday that healthcare.gov is stable and working well, a far cry from last year's frozen computer screens and frustrated customers. Health and human services secretary sylvia burwell smiles while answering a reporters question following her tour of the greater prince william community health center, evergreen terrace site, in manassas,... ( associated press ) health and humanservices secretary syvla m. Burwell gestures while answering the reporters question after her tour on the greater Prince william Community Health Center, Evergreen Terrace site in Manassas, Va. ( associated Press ) health care.gov website in Washington, D.C. ( Associated Press ) In an interview with NBC, Burwell said: "Healthcare.gov works really well now," adding that "in addition to new applicants," 500,000 people were able to log on to healthcare.com this week, and 1 million people have gone "window shopping" to compare insurance plans and prices over the past week. \\
        \end{tabularx}
    \end{promptbox}
    \caption{Example generated summaries from PRIMERA models for a Multi-News+ example. Source documents and ground-truth reference summary are in \Cref{fig:multinews_example}.}
    \label{fig:primera_summaries}
\end{figure*}

\begin{figure*}[]
    \centering
    \begin{promptbox}{Generated Summaries: Summ$^N$ Models}
        \begin{tabularx}{\linewidth}{@{} >{\RaggedRight}p{2.9cm} >{\RaggedRight}X @{}}
        \modelmetrics{MultiXScience}{Comp-Word: 6.86 \newline R-LSum: 0.20 \newline Fact. (P): 2.00 \newline Fact. (U): 0.93 \newline Fluency (P): 3.00 \newline Fluency (U): 0.80} &
        The secretary of health and human services reported that 100,000 people submitted new applications for Obamacare in the first days of the second open enrollment period @cite . The secretary also said that call centers have taken 100,00 calls, another indication of consumer interest. The administration has projected a total of 13 million enrolled for 2015, and some see the administration as trying to lower expectations. \\
        \midrule
        
        \modelmetrics{ConvoSumm}{Comp-Word: 6.04 \newline R-LSum: 0.21 \newline Fact. (P): 3.00 \newline Fact. (U): 0.97 \newline Fluency (P): 3.00 \newline Fluency (U): 0.96} &
        The secretary of health and human services said that 100,000 people submitted new applications for Obamacare in the first days of the second open enrollment period. She also said that a half-million people who already have coverage through the program were able to log into their accounts this time. The numbers offer a stark contrast to the first enrollment period a year ago when healthcare.gov launched with severe technical problems that prevented people from applying. \\
        \midrule

        \modelmetrics{MultiNews+}{Comp-Word: 3.91 \newline R-LSum: 0.24 \newline Fact. (P): 2.00 \newline Fact. (U): 0.44 \newline Fluency (P): 3.00 \newline Fluency (U): 0.76} &
        The second open enrollment period for obamacare's 2015 enrollment period kicked off today, and the administration says it's already seeing a big jump in applications. Health and human services secretary sylvia burwell told meet the press that 100,000 people applied for coverage on the federal website this weekend, up from 23,000 on the first day last year, reports politico. " healthcare.gov works really well now, " said burwell. The administration expects " strong and healthy growth " for 2015, she said, adding that 7 million people have signed up so far, and she expects that number to grow by 2 million more. The congressional budget office has projected a total of 13 million enrolled for 2015.
        \end{tabularx}
    \end{promptbox}
    \caption{Example generated summaries from Summ$^N$ models for a Multi-News+ example. Source documents and ground-truth reference summary are in \Cref{fig:multinews_example}.}
    \label{fig:sumn_summaries}
\end{figure*}

\begin{figure*}
    \centering
    \begin{promptbox}{Generated Summaries: ConvoSumm Models}
        \begin{tabularx}{\linewidth}{@{} >{\RaggedRight}p{4cm} >{\RaggedRight}X @{}}
        \modelmetrics{MultiXScience}{Comp-Word: 8.55 \newline R-LSum: 0.19 \newline Fact. (P): 3.00 \newline Fact. (U): 0.21 \newline Fluency (P): 1.00 \newline Fluency (U): 0.62} &
        @cite @cite , the secretary of health and human services @cites said that 100,000 people submitted new applications for obamacare in the first days of the second open enrollment period. @ccite also said that a half-million people who already have coverage through the program were able to log into their accounts this time. \\
        \midrule
        
        \modelmetrics{ConvoSumm-Reddit}{Comp-Word: 14.61 \newline R-LSum: 0.08 \newline Fact. (P): 1.00 \newline Fact. (U): 0.88 \newline Fluency (P): 2.00 \newline Fluency (U): 0.96} &
        Most commentators say that the website is stable and working well. One commentator says that the number of new applicants is a good sign. Another commentator says the website looks good. \\
        \midrule
        
        \modelmetrics{MultiNews+}{Comp-Word: 2.44 \newline R-LSum: 0.20 \newline Fact. (P): 3.00 \newline Fact. (U): 0.92 \newline Fluency (P): 3.00 \newline Fluency (U): 0.82} &
        Health and human services secretary sylvia burwell says healthcare.gov is stable and working well, a far cry from last year's frozen computer screens and frustrated customers. Burwell told nbc's meet the press that 100,000 people had submitted new applications this weekend via the federal website serving 37 states. That's a big difference from last week, when only a handful of customers managed to enroll on the first day. Bur Well also said that a half-million people who already have coverage through the program were able to log into their accounts this time. There were reports saturday that returning customers had problems, but some of that may have been confusion trying to remember user names and passwords. Administration spokesman aaron albright said sunday he had not seen any indication of problems. Bur well said she expects " strong and healthy growth " for 2015. About 7 million people are signed up, and burwell expects to grow that by 2 million more or so. The congressional budget office has projected a total of 13 million enrolled for 2015, and some see the administration as trying to lower expectations.
        \end{tabularx}
    \end{promptbox}
    \caption{Example generated summaries from ConvoSumm models for a Multi-News+ example. Source documents and ground-truth reference summary are in \Cref{fig:multinews_example}.}
    \label{fig:convosumm_summaries}
\end{figure*}

\begin{figure*}
    \centering
    \begin{promptbox}{Generated Summary: Llama 3.1 Zero-Shot}
        \begin{tabularx}{\linewidth}{@{} >{\RaggedRight}p{2.9cm} >{\RaggedRight}X @{}}
        \modelmetrics{Zero-Shot}{Comp-Word: 3.00 \newline R-LSum: 0.25 \newline Fact. (P): 4.00 \newline Fact. (U): 0.86 \newline Fluency (P): 4.00 \newline Fluency (U): 0.96} &
        As the second sign-up season for healthcare under the Obama administration gears up, officials report significant improvements over last year's struggles. According to Health and Human Services Secretary Sylvia Burwell, approximately 7 million people are currently signed up, with an expectation to grow by 2 million more or so. This year's enrollment progress is a stark contrast to last year's frozen computer screens and frustrated customers. In the first days of the current enrollment period, 100,000 new applications have been submitted via the federal website, healthcare.gov, which serves 37 states. An additional 500,000 existing customers were able to access their accounts, and 1 million individuals have explored insurance plans and prices on the website. President Barack Obama has noted the improvement, stating that "healthcare.gov works really well now." The administration expects "strong and healthy growth" for 2015, with a projected total of 13 million enrolled, surpassing the previous year's enrollment figures. \\
        \end{tabularx}
    \end{promptbox}
    \caption{Example generated summary from the Zero-Shot model for a Multi-News+ example. Source documents and ground-truth reference summary are in \Cref{fig:multinews_example}.}
    \label{fig:zeroshot_summary}
\end{figure*}

\begin{figure*}
    \small
    \centering
    \begin{promptbox}{Generated Summaries: In-Context Learning (ICL) with Llama 3.1 Models}
\begin{tabularx}{\linewidth}{@{} >{\RaggedRight}p{2.9cm} >{\RaggedRight}X @{}}
\modelmetrics{ConvoSumm-Reddit}{Comp-Word: 3.06 \newline R-LSum: 0.31 \newline Fact. (P): 4.00 \newline Fact. (U): 0.83 \newline Fluency (P): 5.00 \newline Fluency (U): 0.96} &
As the second sign-up season for healthcare under the Obama administration gears up, officials are reporting significant improvements in the stability and functionality of healthcare.gov. Health and Human Services Secretary Sylvia Burwell has stated that the website is now working well, with over 100,000 new applications submitted in the first weekend, compared to just a handful of customers who managed to enroll last year. Additionally, approximately 500,000 customers were able to log into their existing accounts, and over 1 million people have begun "window shopping" to compare insurance plans and prices. These numbers are in stark contrast to the technical difficulties and frustrations experienced during the first enrollment period, and suggest a strong and healthy growth for the program in 2015, with the Congressional Budget Office projecting a total of 13 million enrolled. President Barack Obama has also noted the improvements, stating that healthcare.gov now works "really well." \\
\midrule

\modelmetrics{Multi-XScience}{Comp-Word: 2.45 \newline R-LSum: 0.27 \newline Fact. (P): 3.00 \newline Fact. (U): 0.90 \newline Fluency (P): 4.00 \newline Fluency (U): 0.96} &
As the second sign-up season for healthcare under the Obama administration gears up, officials are reporting significant improvements in the stability and functionality of healthcare.gov. Health and Human Services Secretary Sylvia Burwell has stated that the website is now working well, in stark contrast to last year's technical issues that plagued enrollment efforts. According to Burwell, over 100,000 new applications have been submitted via the federal website, serving 37 states, since the beginning of the second open enrollment period. This is a significant increase from last year, when only a handful of customers managed to enroll on the first day. \par Additionally, over 500,000 existing customers have been able to log into their accounts, and 1 million people have gone "window shopping" to compare insurance plans and prices over the past week. The numbers suggest a high level of consumer interest and a continued growth in enrollment. The Congressional Budget Office has projected a total of 13 million enrolled for 2015, and Burwell expects to reach this goal or come close. President Barack Obama has also noted the improvements, stating that "healthcare.gov works really well now." \\
\midrule

\modelmetrics{MultiNews+}{Comp-Word: 2.49 \newline R-LSum: 0.25 \newline Fact. (P): 3.00 \newline Fact. (U): 0.94 \newline Fluency (P): 4.00 \newline Fluency (U): 0.96} &
As the second sign-up season for the Affordable Care Act (ACA) kicks off, the Obama administration is reporting significant improvements in the healthcare.gov website. Health and Human Services Secretary Sylvia Burwell has stated that the website is stable and working well, in stark contrast to last year's technical issues that frustrated customers. Despite some initial concerns about returning customers experiencing problems logging into their existing accounts, Burwell has reported that 500,000 people were able to access their accounts on the website. \par In the first weekend of the new open enrollment period, 100,000 new applications were submitted through healthcare.gov, with an additional 100,000 calls made to call centers. Additionally, 1 million people have gone "window shopping" to compare insurance plans and prices over the past week. These numbers represent a significant increase from last year's enrollment period, where only a handful of customers managed to enroll on the first day. Burwell has expressed confidence in the website's ability to handle the increased traffic, stating that she expects "strong and healthy growth" for 2015, with 13 million people projected to enroll through the website. \\ 
\end{tabularx}
    \end{promptbox}
    \caption{Example generated summaries from In-Context Learning (ICL) models for a Multi-News+ example. Source documents and ground-truth reference summary are in \Cref{fig:multinews_example}.}
    \label{fig:icl_summaries}
\end{figure*}